\pdfoutput=1
\documentclass[11pt]{article}

\usepackage{EACL2023}

\usepackage{times}
\usepackage{latexsym}
\usepackage{xcolor}
\usepackage{soul}
\usepackage{hyperref}
\usepackage{amsmath}
\usepackage{graphicx}
\usepackage{bbm}
\usepackage{booktabs}
\usepackage{multirow}
\usepackage{algpseudocode}
\usepackage{algorithm}
\usepackage{amssymb}
\usepackage[normalem]{ulem}

\usepackage[T1]{fontenc}
\usepackage[utf8]{inputenc}

\usepackage{microtype}

\usepackage{inconsolata}
\usepackage{xcolor}
\definecolor{mycolor1}{HTML}{0000AA}
\definecolor{mycolor2}{HTML}{BB0000}
\definecolor{mycolor3}{HTML}{002200}
\let\svthefootnote\thefootnote
\newcommand\freefootnote[1]{%
  \let\thefootnote\relax%
  \footnotetext{#1}%
  \let\thefootnote\svthefootnote%
}
\newcommand{\methodname}{\textsc{LongEval}}

\newcommand{\titlestr}{\methodname: Guidelines for Human Evaluation of \\  Faithfulness in Long-form Summarization}

\title{\titlestr}

\author{Kalpesh Krishna$^{\spadesuit\,*}$ \quad Erin Bransom$^{\diamondsuit}$ \quad Bailey Kuehl$^\diamondsuit$ \\  {\bf Mohit Iyyer}$^\spadesuit$ \quad {\bf Pradeep Dasigi}$^{\diamondsuit}$  \quad {\bf Arman Cohan}$^{\diamondsuit\heartsuit}$ \quad {\bf Kyle Lo}$^{\diamondsuit}$ \vspace{8pt}\\
$^\spadesuit$University of Massachusetts Amherst, $^\diamondsuit$Allen Institute for AI, $^\heartsuit$Yale University \\ \texttt{\small\{kalpesh,miyyer\}@cs.umass.edu}\\ \texttt{\small\{erinbransom,baileyk,pradeepd,armanc,kylel\}@allenai.org} }

\newcommand{\numsurveypapers}{162} 
\newcommand{\namedref}[2]{\hyperref[#2]{#1~\ref*{#2}}}
\newcommand{\sectionref}[1]{\namedref{Section}{#1}}
\newcommand{\tableref}[1]{\namedref{Table}{#1}}
\newcommand{\figureref}[1]{\namedref{Figure}{#1}}
\newcommand{\appendixref}[1]{\namedref{Appendix}{#1}}
\newcommand{\algorithmref}[1]{\namedref{Algorithm}{#1}}

\newcommand{\coarse}{\textsc{coarse}}
\newcommand{\fine}{\textsc{fine}}

\begin{document}
\maketitle
\begin{abstract}

While human evaluation remains best practice for accurately judging the faithfulness of automatically-generated summaries, few solutions exist to address the increased difficulty and workload when evaluating \emph{long-form} summaries.
Through a survey of \numsurveypapers~papers on long-form summarization, we first shed light on current human evaluation practices surrounding long-form summaries. We find that 73\% of these papers do not perform any human evaluation on model-generated summaries, while other works face new difficulties that manifest when dealing with long documents (e.g., low inter-annotator agreement). Motivated by our survey,
we present \methodname, a set of guidelines for human evaluation of faithfulness in long-form summaries that addresses the following challenges: 
(1) How can we achieve high inter-annotator agreement on faithfulness scores?
(2) How can we minimize annotator workload while maintaining accurate faithfulness scores?
and (3) Do humans benefit from automated alignment between summary and source snippets? We deploy \methodname\ in annotation studies on two long-form summarization datasets in different domains (SQuALITY and PubMed), and we find that switching to a finer granularity of judgment (e.g., clause-level) reduces inter-annotator variance in faithfulness scores (e.g., std-dev from 18.5 to 6.8).
We also show that scores from a \emph{partial} annotation of fine-grained units highly correlates with scores from a full annotation workload (0.89 Kendall's $\tau$ using 50\% judgments). 
%Finally, we illustrate when and how automated summary-source alignment does (or does not) help human evaluation. 
We release our human judgments, annotation templates, and our software for future research.\footnote{\url{https://github.com/martiansideofthemoon/longeval-summarization} \\ 
*Work done in an AI2 internship, author contributions \hyperref[sec:contrib]{here}.
}
\end{abstract}

\section{Introduction}

%\kkcomment{todo: section transitions to reflect updated survey, shorten to 8 pages, code / data / checklist}

% \kkcomment{todo - transitions / 2 paragraphs in linking section, figure}
% \klcomment{Cut first sentence; discussion of ROUGE seems off topic. go straight into discussion of human-eval.}
% While ROUGE~\citep{lin-2004-rouge} has been the dominant paradigm for evaluating model-generated summaries, it correlates poorly with human judgments of summary quality~\citep{cohan-goharian-2016-revisiting,goyal2022news}.\micomment{there are def more papers you can cite pre 2016 about the failings of ROUGE} 
% \micomment{this sentence doesn't really flow with the previous one. can you connect it to ROUGE somehow, maybe with a comparison of human eval vs rouge?} 
%
%Recent works have managed to assemble large collections of human evaluations of model-generated summaries~\citep{fabbri2021summeval,pagnoni-etal-2021-understanding}.
Human judgments are considered the gold standard for evaluating model-generated summaries~\citep{kryscinski-etal-2019-neural,fabbri2021summeval} and generated text more broadly~\citep{celikyilmaz2020evaluation}. Unfortunately, human evaluation tends to be labor-intensive, expensive to scale, and difficult to design. This is problematic as a large number of judged examples is needed to draw statistically significant conclusions about system performances~\citep{wei-jia-2021-statistical} or correlations between human judgments and automatic metrics~\citep{deutsch2021statistical}. Human evaluation is especially challenging when \emph{long} sequences of generated text need to be evaluated, due to the inherent subjectivity in the task~\citep{karpinska-etal-2021-perils, clark-etal-2021-thats, krishna-etal-2021-hurdles, goyal2022snac}.

\begin{figure*}[t!]
    \centering
    \includegraphics[width=0.99\textwidth]{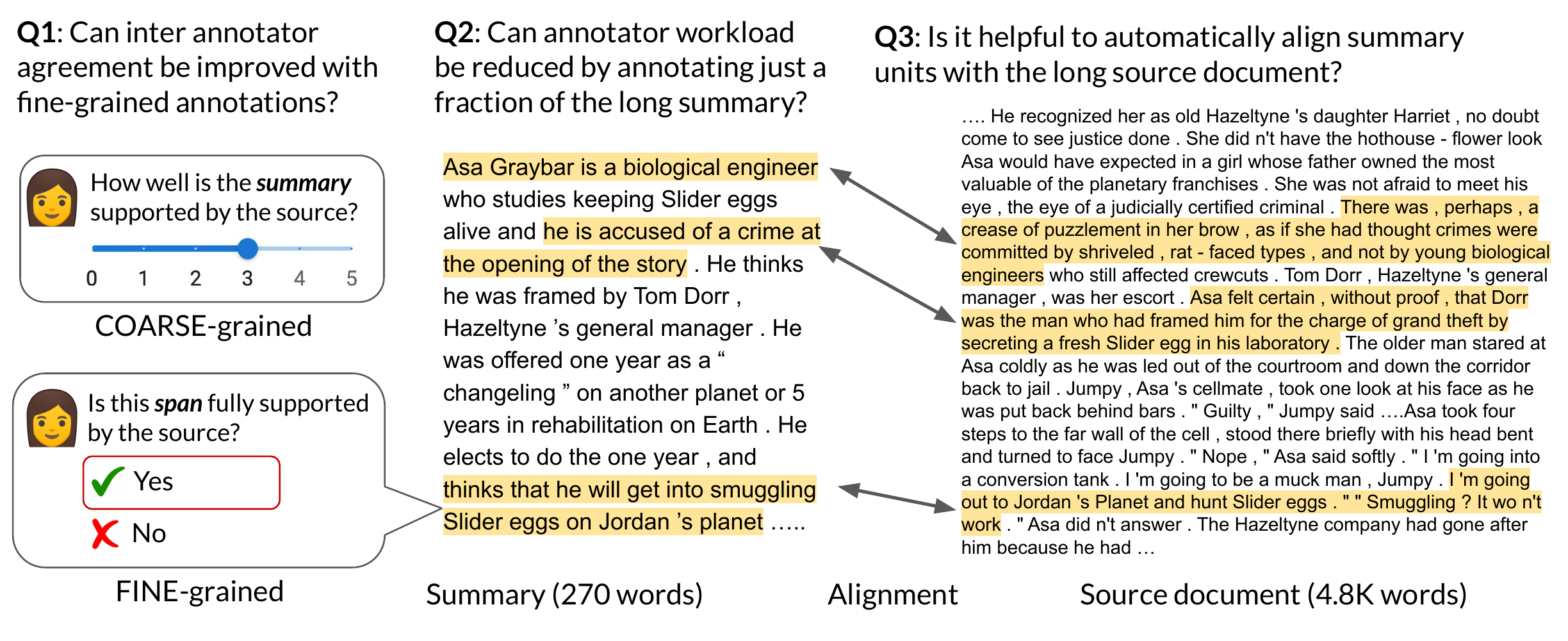}
    \caption{Overview of research questions considered in \methodname.  Example summary taken from SQuALITY.}
    \label{fig:skimeval-diagram}
\end{figure*}

To better understand the challenges of human evaluation on long-form summaries (150 words or longer), we first conduct a comprehensive survey of \numsurveypapers{} publications and preprints on long-form summarization (\sectionref{sec:survey}). We find that 119 papers (73\%) do not perform human evaluation on long-form summaries, while the remaining papers deviate significantly from suggested best practices for reproducibility~\citep{gehrmann2022repairing}. Current human evaluation setups lack standardization in their design decisions (such as annotation granularity), some of which can significantly impact inter-annotator agreement (\sectionref{sec:fine-vs-coarse-annotations}). Finally, 20 papers explicitly mention human evaluation is expensive, difficult, and time-consuming due to the long length of summaries and source documents.

%for instance, in one such surveyed paper~\citep{wang2022squality} the average difference between the maximum / minimum faithfulness rating across three annotators is 42.5 on a scale of 100!
%, annotation budget, and the overall validity of the human evaluation, especially when evaluating long generations. \klcomment{this statement - ``decisions can impact things'' - is a bit weak. can we provide examples of the consequences of bad design decisions from these papers? i want to know ``which'' decisions were impactful and in what negative way, to set up motivation for us investigating the 3 focus questions of our work.}\kkcomment{todo}

To move towards a more consistent and efficient human evaluation, we present \methodname, a set of guidelines for human evaluation of faithfulness in long-form summarization (\sectionref{sec:longeval}). We empirically evaluate \methodname\ using human annotation studies on two long-form summarization datasets:  SQuALITY~\citep{wang2022squality} and PubMed~\citep{cohan-etal-2018-discourse}. We provide an overview of our main research questions and findings in \figureref{fig:skimeval-diagram} and enumerate them here:

\vspace{0.1in}

\noindent \textbf{RQ1}: \emph{Can inter-annotator agreement be improved while evaluating faithfulness of long-form summaries via fine-grained annotations?} \\

\vspace{-0.1in}

\noindent \textbf{Finding}: Annotating faithfulness of individual summary clauses and aggregating them leads to significantly higher inter-annotator agreement, compared to the dominant paradigm of evaluating whole summaries at once via Likert ratings (std-dev 18.5 to 6.8 on SQuALITY).

\vspace{0.2in}

\noindent \textbf{RQ2}: \emph{Can we reduce annotator workload by partially annotating a long summary while maintaining accurate faithfulness scores?}\\

\vspace{-0.1in}

\noindent \textbf{Finding}: Despite annotating a fraction of summary clauses, 
faithfulness scores under a reduced workload maintain high correlation with those from a full workload (0.89 Kendall's $\tau$ at 50\% workload).

\vspace{0.2in}

\noindent \textbf{RQ3}: \emph{Do humans benefit from automatically aligning summary units to relevant sentences in the source document?}\\

\vspace{-0.1in}

\noindent \textbf{Finding}: Unlike suggestions in prior work on short-form summarization~\citep{hardy-etal-2019-highres,kryscinski-etal-2020-evaluating}, aligning parts of the summary to source document is only useful when the summary is highly extractive or mostly correct.

\vspace{0.2in}

\noindent Overall, our contributions are:

\noindent \textbf{(1)} a \numsurveypapers-paper survey of current human evaluation practices in long-form summarization;

\noindent \textbf{(2)} \methodname, a set of three guidelines for evaluating faithfulness in long-form summarization; 

\noindent \textbf{(3)} an empirical validation of \methodname\ guidelines on two long-form summarization datasets in different domains (SQuALITY and PubMed);

\noindent \textbf{(4)} A dataset with 3-way fine-grained human faithfulness judgments for 120 SQuALITY \& PubMed summaries annotated using \methodname{} which can be used for benchmarking automatic metrics.

 We open-source our human evaluation data, annotation interface, and code for future research.\footnotemark[1]

%\accomment{Maybe another way to go about this is to first describe the main components of the framework. Then describe the findings. Right now you have the main findings in each bullet point, then describe the principle that lead to that finding. Esp for point \#3 just reading the bullet point can be confusing because the reader doesn't know yet why we are talking about highlighting. }

%Instead of obtaining a single judgment of the whole summary at once (via Likert scale), we suggest obtaining binary judgments of \emph{individual units} like phrases in the generated summary. Full summary scores are obtained by averaging binary judgments across the summary. This makes annotation more well-defined, lower variance, and provides finer-grained error annotations.

% Finally, to assist human evaluators navigate through a long source document, we highlight the most relevant source document sentence for every summary clause.

% Annotating every individual clause can be prohibitively expensive for long summaries. Instead, we annotate a random subset of clauses (25\% clauses in our experiments) in the summary to estimate the overall faithfulness.\kkcomment{make more precise after all annotations done}

\section{Survey of human evaluation practices}
\label{sec:survey}

Before discussing \methodname, we first attempt to understand current human evaluation practices in long-form summarization through a comprehensive survey of 162 papers. Our survey reveals several concerning trends: absence of human evaluation, non-reproducible experimental setups, lack of standardization, and complaints of long summaries being challenging and expensive to evaluate. These results show an urgent need to develop more efficient and standardized human evaluation protocols.

\vspace{0.1in}

%\pdcomment{you may want to summarize your main takeaways from the summary here before you go into the details. Currently it is unclear how this survey is related to the decisions made for SkimEval.\micomment{agreed}}\kkcomment{bullet list of main takeaways --- connect it well to Section 3.. some ideas - lack of best practices, variance stats reported, complaints of it being expensive / time-consuming, long source documents means navigating it is hard}

\noindent \emph{Selection of papers:} We consider existing summarization datasets with an average summary length of at least 150 words, which includes several popular datasets like arXiv~\citep{cohan-etal-2018-discourse}, BillSum~\citep{kornilova-eidelman-2019-billsum} and MultiNews~\citep{fabbri-etal-2019-multi}; see \tableref{tab:dataset-survey} for a full list. For our survey, we select all papers that evaluated summarization models using at least one of these datasets.\footnote{We exclude five papers which used long-form summarization data for pre-training only, like~\citet{wei2022finetuned}.} All of these papers were published between June 2018 and September 2022, after the first long-form summarization datasets were released (PubMed / arXiv). Most of the \numsurveypapers\ surveyed papers were published in major NLP/ML venues, but we also include newer preprints from 2022.

\vspace{0.1in}

\noindent \textbf{Long-form summaries are rarely evaluated by humans.} We find that 101 out of 162 papers (62\%) do not perform any human evaluation. 17 papers (11\%) only perform human evaluation on short summaries (datasets like XSUM,~\citealp{narayan-etal-2018-dont}), for which human evaluation is much easier.

\vspace{0.1in}

\noindent \textbf{Human evaluation studies of long-form summaries are not reproducible.} We further analyze the 44 papers performing human evaluation of long-form summaries to observe how often they follow reproducible practices from~\citet{gehrmann2022repairing}. Overall, we find that most studies do not follow these guidelines. Only 2 of the 44 papers release their raw human annotation data for further analysis. Only 9 papers provide details of their annotator instructions or interface, and just 12 papers perform any kind of statistical analysis, despite most papers annotating less than 50 summaries. While 33 papers report using multiple annotators per summary, only 12 report inter-annotator agreement. Finally, just 14 papers conduct human evaluation on more than one dataset (more statistics in \appendixref{sec:additional-survey-stats}).

\vspace{0.1in}

\noindent \textbf{Existing human evaluation setups lack standardization.} In \tableref{tab:results-survey}, we catalog the wide spectrum of human evaluation setups in the surveyed papers. 37 papers collect judgments of the full-length summary at once (``\coarse-grained''), while 6 papers collect judgments at a finer granularity such as sentences or entities (``\fine-grained''). Even within a granularity, setups differ: Likert-scale (24 papers), A/B testing (13 papers), binary per-sentence labels (4 papers) are the dominant protocols. In \sectionref{sec:fine-vs-coarse-annotations}, we will see that this design decision is critical since \coarse\ annotations have much lower inter-annotator agreement than \fine.\footnote{Besides granularity, we also observe a large spectrum of annotator qualifications in our survey, ranging from MTurkers to expert graduates (\appendixref{sec:additional-survey-stats}). Since non-experts are known to be unsuitable for this task~\citep{gillick2010non,fabbri2021summeval}, we use experts in our work (\appendixref{sec:human-evaluation-details}).}

\vspace{0.1in}

\noindent \textbf{Human evaluation of long-form summaries is challenging and expensive.} Several of the surveyed papers discuss challenges in human evaluation of long-form summaries. 13 papers mention that expert annotators are necessary for human evaluation of long-form summaries, especially in technical domains like PubMed. 20 papers report that human evaluation of long-form summarization was \emph{time-consuming}, \emph{challenging}, and \emph{expensive}, primarily due to the long length of the summary and source document. To tackle the issue of high annotator workload, we propose a partial annotation method in \sectionref{sec:partial-annotation} and report high correlation to a full workload. Additionally, in \sectionref{sec:highlight-hints} we investigate the usefulness of highlighting sentences to help annotators navigate the long source document. While this has been advocated for in short-form summary evaluation~\citep{hardy-etal-2019-highres,kryscinski-etal-2020-evaluating} and used in 3 surveyed long-form papers, we find that it is only helpful when summaries are mostly correct and extractive.

\begin{table}[t]
\small
\begin{center}
\begin{tabular}{ lrrr } 
 \toprule
 Dataset & $|$source$|$ & $|$summary$|$ & papers \\
 & (words) & (words) & \\
 \midrule
 PubMed~\shortcite{cohan-etal-2018-discourse} & 3092 & 205 & 59 \\
 arXiv~\shortcite{cohan-etal-2018-discourse} & 5906 & 163  & 55 \\
 BillSum~\shortcite{kornilova-eidelman-2019-billsum} & 1284 & 174  & 19 \\
 MultiNews~\shortcite{fabbri-etal-2019-multi} & 2103 & 263  & 54 \\
 GovReport~\shortcite{huang-etal-2021-efficient} & 7551 & 547  & 16 \\
 BookSum~\shortcite{kryscinski2021booksum} & 5102 & 505  & 4 \\
 SummScreen~\shortcite{chen-etal-2022-summscreen} & 6965 & 227  & 11 \\
 SQuALITY~\shortcite{wang2022squality} & 5194 & 227  & 1\\
\bottomrule
\end{tabular}
\end{center}
\caption{List of long-form summarization datasets considered in our survey along with average source document and summary lengths. Each dataset considered has at least 150 word summaries on average.}

\label{tab:dataset-survey}
\end{table}

\begin{table}[t]
\small
\begin{center}
\begin{tabular}{ lrr } 
 \toprule
 Type of human evaluation & \# papers & \% papers \\
 \midrule
None & 101 & 62\% \\
Short-form summaries only & 17 & 11\% \\
\midrule
Likert-scale \coarse-grained & 24 & 15\% \\
A/B testing \coarse-grained & 13 & 8\% \\
Extrinsic evaluation & 1 & 1\% \\
Binary per sentence \fine-grained & 4 & 2\% \\
QA-based \fine-grained & 2 & 1\% \\
\bottomrule
\end{tabular}
\end{center}
\caption{Human evaluation setup in \numsurveypapers\ summarization papers that evaluate long-form summaries. 73\% of the papers do not evaluate long-form summaries with humans, while others vary significantly in their setups.}
\label{tab:results-survey}
\end{table}

%\noindent \textbf{Variance of Human Annotations}: Prior work has shown that holistic evaluation of long-form generation is challenging~\citep{clark-etal-2021-thats,krishna-etal-2021-hurdles} and high variance across annotators~\citep{karpinska-etal-2021-perils,goyal2022snac}. How much of an issue is\micomment{wording is weird, do you mean to say something like ``how large/high is inter-annotator variance''} inter-annotator variance in our surveyed papers? Unfortunately, only 3 out of 23 papers performing Likert-scale\footnote{\kkcomment{Discuss A/B test too in footnote}} holistic evaluation report inter-annotator agreement in any form (often vaguely). None of the papers release their raw human annotations, consistent with a recent survey~\citep{gehrmann2022repairing} on the lack of reproducibility in human evaluation of text generation. \kkcomment{can reword previous paragraph due to new pitch, overall there's no clear consensus on best practices}
\section{The \methodname\ guidelines for faithfulness human evaluation}
\label{sec:longeval}

In \sectionref{sec:survey}, we report several concerning issues with current human evaluation practices in long-form summarization. To move towards more efficient, reproducible and standardized protocols for human evaluation,  we develop the \methodname\ guidelines (\sectionref{sec:fine-vs-coarse-annotations}-\ref{sec:highlight-hints}, see \figureref{fig:skimeval-diagram} for an overview). We focus on human evaluation of \emph{faithfulness}, which \citet{wang2022squality} define as:

\begin{quote}
{
\footnotesize``\emph{Checking the factual errors in the summary, where a factual error is a
statement that contradicts the source document, or is not directly stated, heavily implied, or logically entailed by the
source document}''
}
\end{quote}

% \vspace{0.1in}

We conduct human annotation studies to empirically motivate \methodname. Our experiments are on two long-form summarization \textbf{datasets} spanning diverse domains and levels of abstractiveness: 

\vspace{0.1in}

\noindent (1) \textbf{SQuALITY} \citep{wang2022squality} is a summarization dataset in the literary domain (avg. summary length of 227 words) where summaries  describe the plots of English science fiction stories. SQuALITY is highly abstractive: on average just 16\% of bigrams in the summary are present in the source document. We closely follow the human evaluation setup in~\citet{wang2022squality}, and use BART~\citep{lewis-etal-2020-bart} and BART-DPR~\citep{karpukhin-etal-2020-dense} as our summarization models along with human-written summaries.

\vspace{0.1in}

\noindent (2) \textbf{PubMed} \citep{cohan-etal-2018-discourse} is a  summarization dataset in the scientific domain  (avg. summary length of 205 words) that pairs English biomedical articles from PubMed\footnote{\url{https://pubmed.ncbi.nlm.nih.gov/}} with their abstracts as summaries. Compared to SQuALITY, PubMed is more extractive: 54\% of summary bigrams are present in the source. We use BigBird-PEGASUS-large~\citep{zaheer2020big} and LongT5-large~\citep{guo-etal-2022-longt5} as our summarization models,\footnote{LongT5 is the best publicly available PubMed summarizer. BigBird is a popular long-form summarization baseline.} along with human written summaries. By default, LongT5 / BigBird were highly extractive compared to human-written PubMed summaries (87\% / 74\% vs 54\% bigram overlap with source). Hence, for half the generations we block 6-grams from being copied from the source,\footnote{Reducing extractiveness / copying is also a suggestion for fair-use of copyrighted work~\citep{harvardcopy16, marylandcopy20}.} reducing extractiveness to $\sim$54\%. We call this setting ``PubMed-ngram-block''.

%\noindent \textbf{Obtaining baseline human annotations}: To conduct further analysis, we contact the authors of each of the 23 papers conducting Likert-scale evaluations of the whole summary, requesting their raw human evaluation data. Only~\citet{wang2022squality} got back to us positively,\footnote{ Each of these had high inter-annotator variance of 1.9-2.2 std-dev on a 5-point scale.} providing us with the human annotations for SQuALITY summaries. \kkcomment{can skip most of previous paragraph in new pitch}

\begin{figure*}[t!]
    \centering
    \includegraphics[width=0.48\textwidth]{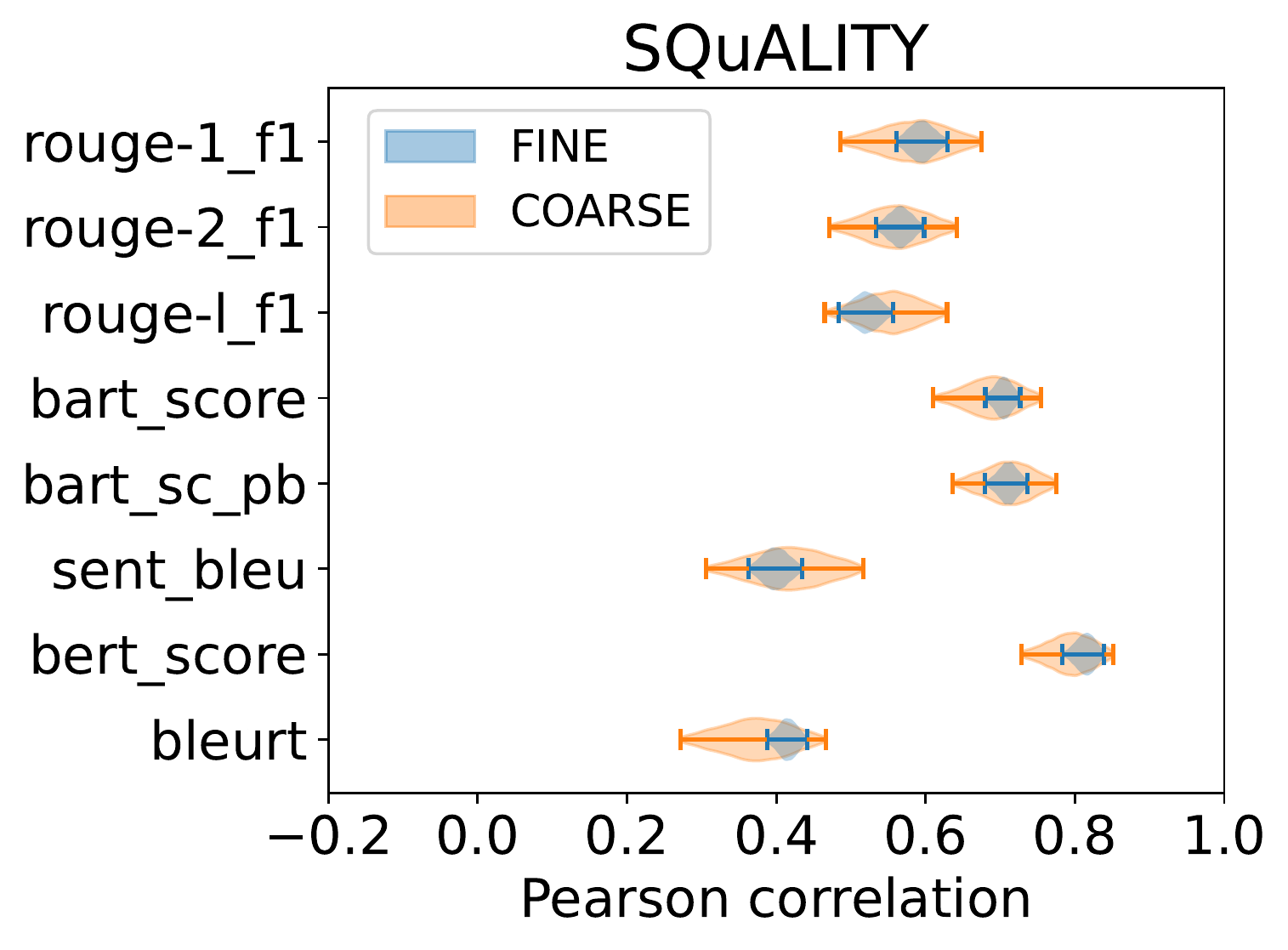}
    \includegraphics[width=0.48\textwidth]{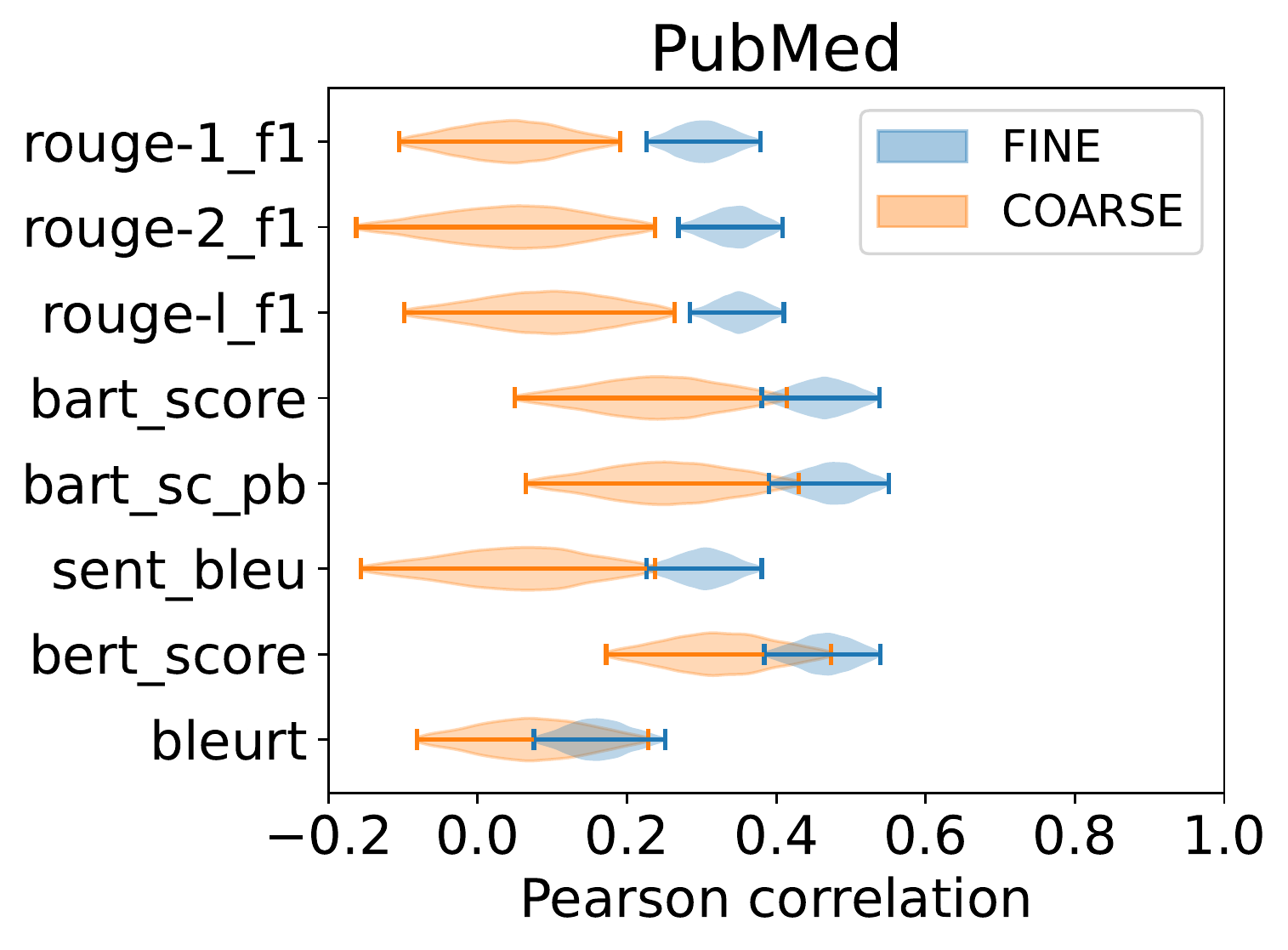}
    \caption{
    % % the 95\% confidence intervals of the Pearson correlation of various automatic evaluation metrics against \fine-grained and \coarse-grained human evaluation data. 
    95\% confidence intervals of Pearson correlations between various automatic evaluation metrics and using human evaluation data collected with \fine\ (blue) and \coarse\ (orange) annotation methods. In both datasets, \fine\ annotations lead to much narrower CIs than \coarse\ annotations. See \appendixref{sec:metric-correlations-kt} for plot with Kendall's Tau.}
    \label{fig:confidence-intervals-metrics}
\end{figure*}

\begin{figure*}[t!]
    \centering
    \includegraphics[width=0.325\textwidth]{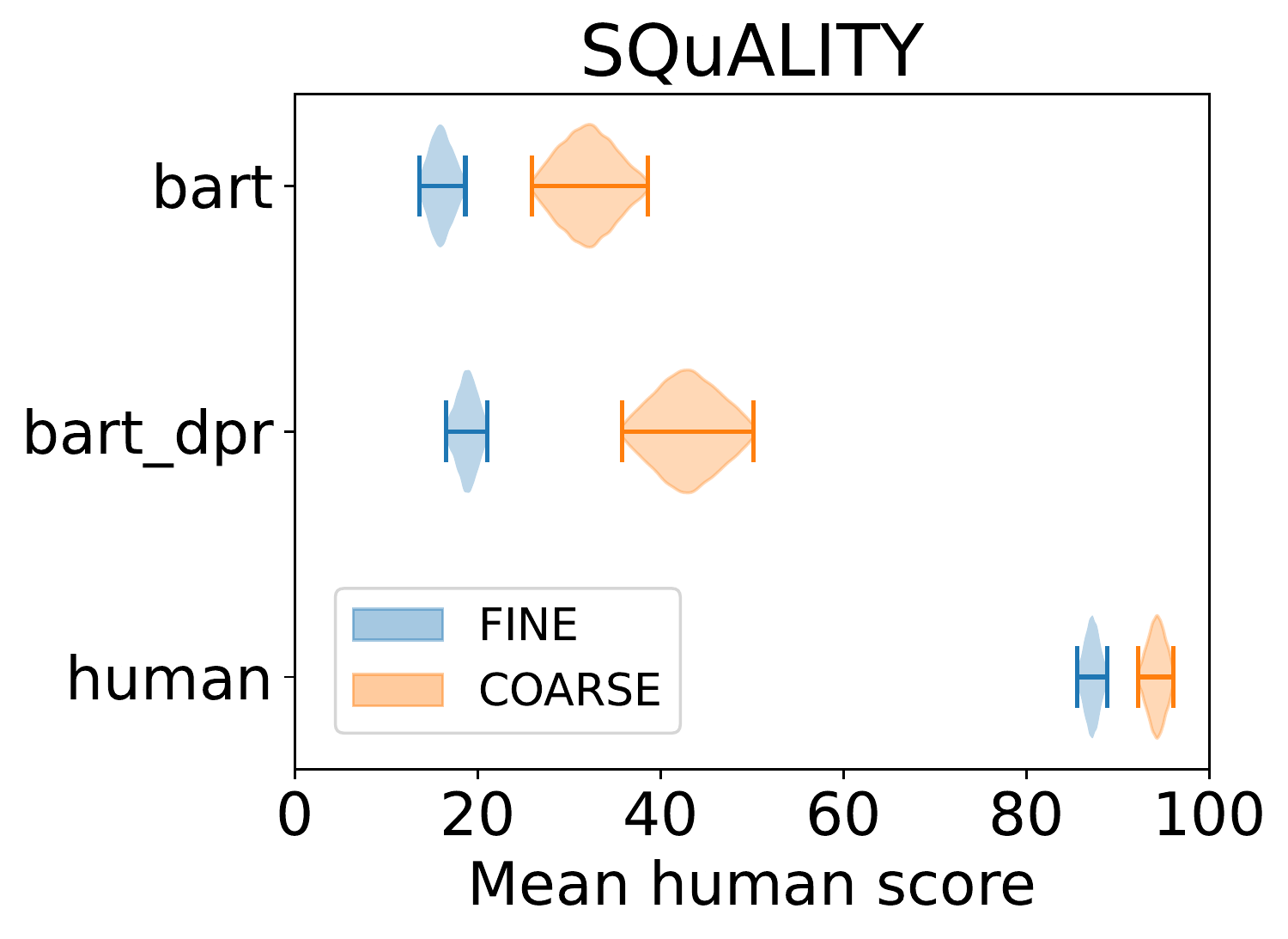}
    \includegraphics[width=0.325\textwidth]{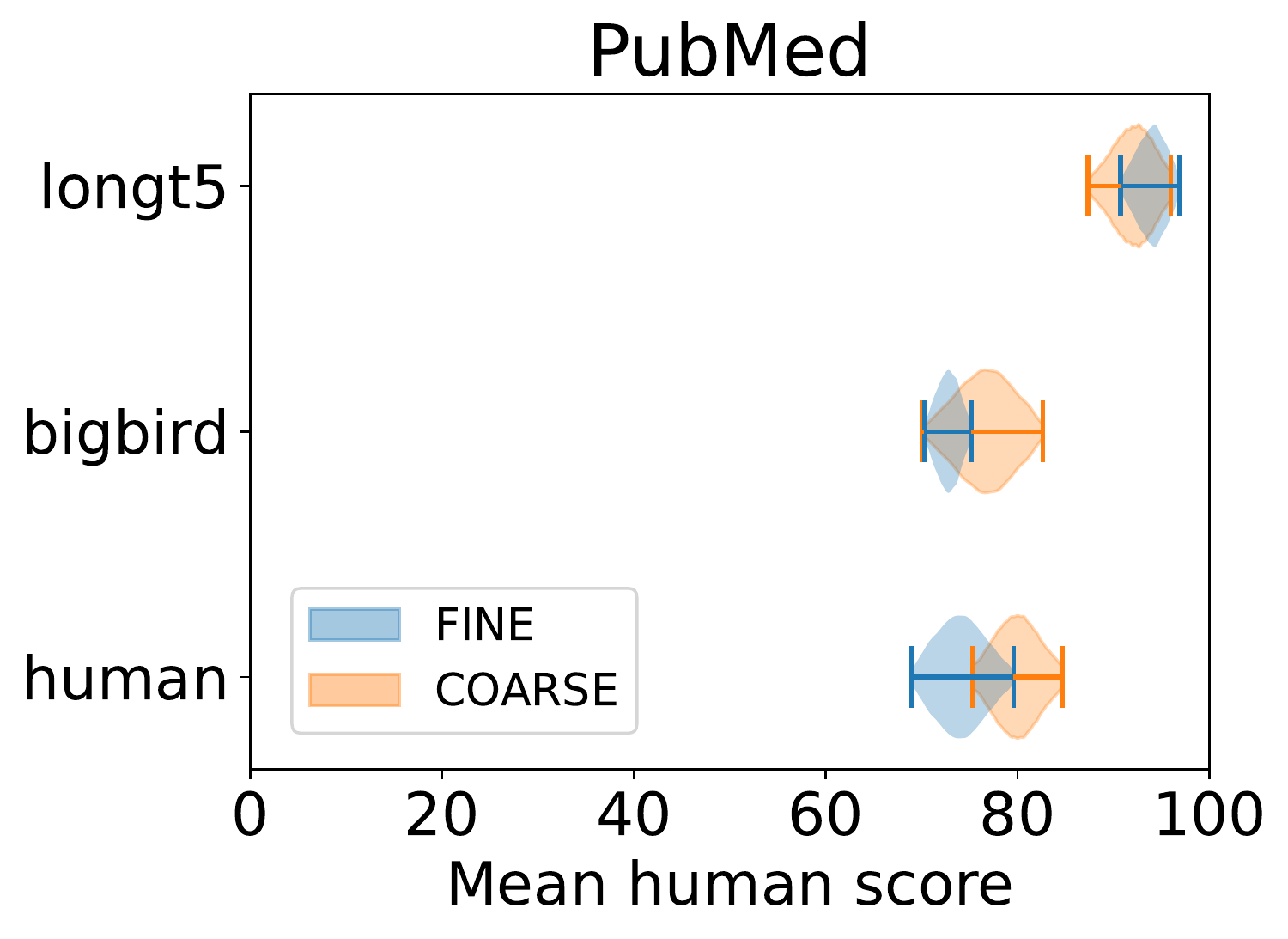}
    \includegraphics[width=0.325\textwidth]{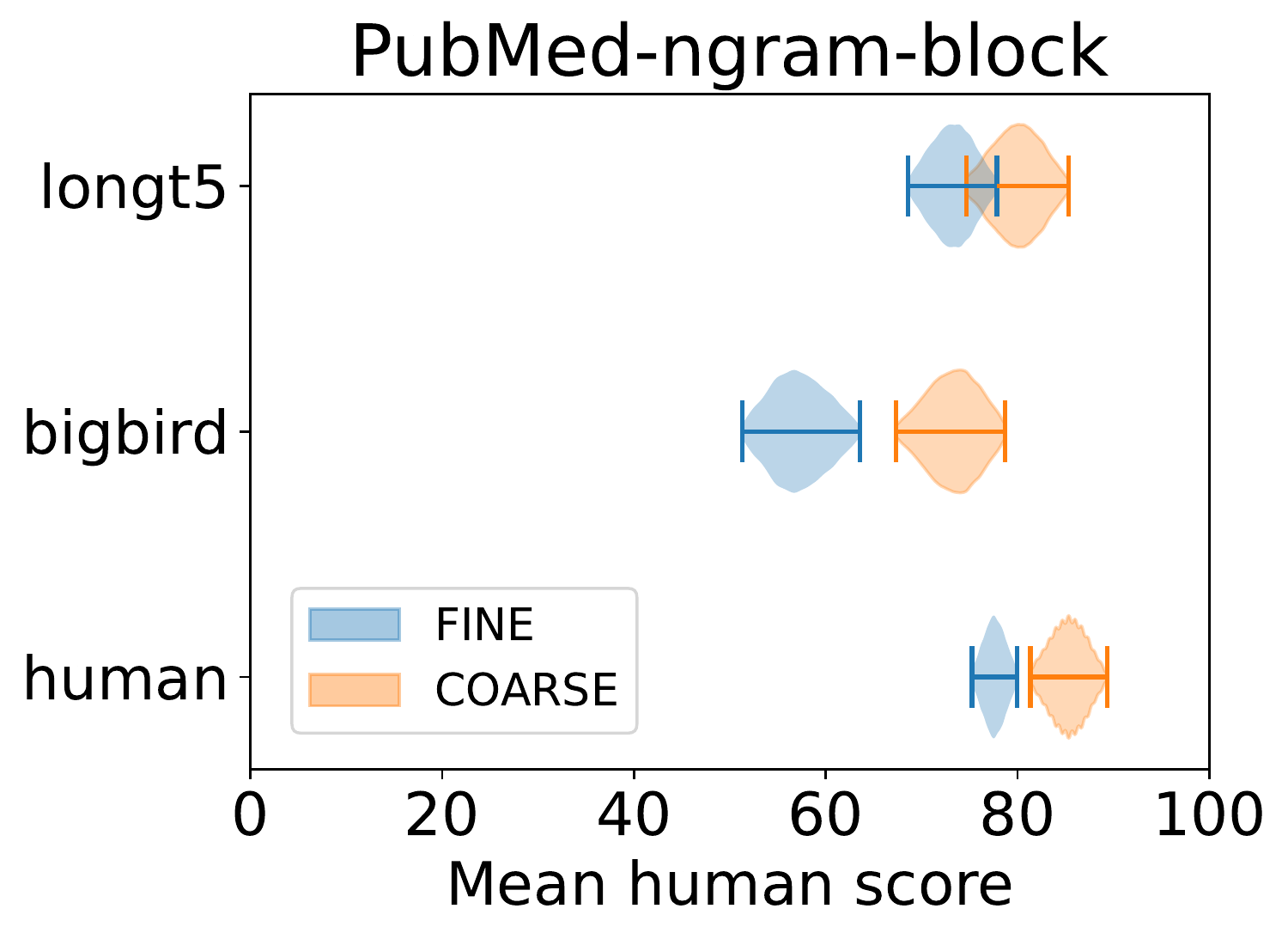}
    \caption{
    95\% confidence intervals of estimated model performances using \fine\ (blue) and \coarse\ (orange) annotation methods.
    Intervals calculated using bootstrap resampling across annotators (\appendixref{sec:bootstrap}). While both annotation granularities lead to similar relative ordering of systems, \fine\ annotations have narrower confidence intervals. The higher LongT5 score vs human in PubMed is due to highly extractive LongT5 summaries (\sectionref{sec:longeval}).}
    \label{fig:confidence-intervals-means}
\end{figure*}

\subsection{RQ1: Does inter-annotator agreement improve using fine-grained annotations?}
\label{sec:fine-vs-coarse-annotations}

In \sectionref{sec:survey}, we found that the dominant paradigm in literature (37 out of 44 papers) is to evaluate the whole summary at once (``\coarse''-grained, \figureref{fig:skimeval-diagram} top left). 6 papers instead obtain fine-grained annotations for individual units (e.g., sentences) and average them (\fine, \figureref{fig:skimeval-diagram} top right). Intuitively, \fine\ annotation has many advantages for longer summaries --- it is less subjective than \coarse, since shorter spans needs to be judged rather than a long summary, and it helps localize model errors. However, the distinction between \coarse\ and \fine\ is never justified in literature, and inter-annotator agreement is rarely reported to understand the task subjectivity in each setup. To better understand the tradeoff, in this section we conduct human evaluations annotating the same set of summaries using these two different protocols.

\vspace{0.1in}

\noindent \textbf{Task formulation}: Let $F_\text{summ}$ denote the faithfulness score of a summary. For \coarse, $k$-point Likert scale ratings are obtained for the summary ($F_\text{summ} \in \{0, 1 ... k\}$), based on the faithfulness definition provided earlier. For \fine, we collect binary judgments of individual units in the summary and average them,
\begin{align*}
    F_{\text{summ}} &= \frac{1}{|\mathcal{C}_{\text{summ}}|} \sum_{c \in \mathcal{C}_{\text{summ}}} F_c,~F_c \in \{0, 1\}
\end{align*}
where $\mathcal{C}_{\text{summ}}$ is a set of units in the summary and $F_c$ is the faithfulness judgment for the unit $c$. In both protocols, the faithfulness score of a system is defined as $\frac{1}{|\mathcal{S}|} \sum_{\text{summ} \in \mathcal{S}} F_\text{summ}$ where  $\mathcal{S}$ is the set of summaries generated by the system.\footnote{We assume all summary units get an equal weight. However, some units may be more important than others, we discuss this in the Limitations section.}

While sentences are a popular granularity for \fine\ (4 of the 6 surveyed papers), we found that summary sentences in both datasets were overloaded with information. Hence, we segment sentences on conjunctions and punctuation to obtain more atomic units as $\mathcal{C}_{\text{summ}}$. These units are often clauses,\footnote{An even finer granularity is entities / numbers. We avoid this due to prohibitive annotation cost on long summaries.} similar to summary content units (SCUs) in Pyramid~\citep{nenkova-passonneau-2004-evaluating}.

\vspace{0.1in}

\noindent \textbf{Collecting \coarse\ annotations}: For SQuALITY, we re-use the annotations provided by~\citet{wang2022squality} for faithfulness assessments. In their data, three annotators give each summary a 1-100 direct assessment rating~\citep{bojar-etal-2016-results}. Annotators with experience in professional copyrighting and editing were hired on Upwork,\footnote{\url{https://www.upwork.com/}}  and these annotators were also involved in the creation of SQuALITY. Unfortunately, none of the surveyed papers that reported human evaluation results on PubMed released their raw human annotations.\footnote{In our email correspondence with authors of these works, they mentioned losing access or compliance issues as reasons for not sharing human evaluations. We received some examples from \citet{guo2021automated} and \citet{ju-etal-2021-leveraging-information} for reference.} Hence, we collect our own \coarse\ evaluations on PubMed summaries on Upwork, using freelancers with professional experience reading and writing research papers (details in \appendixref{sec:coarse-grained-human-evaluation}). We collect 3 annotations per summary and use a 5-point Likert scale, the most common choice for \coarse\ assessment in our survey (18 out of 38 papers). In total, 120 summaries are evaluated.

\vspace{0.1in}

\begin{table}[t]
\begin{center}
\begin{tabular}{ lrr } 
 \toprule
Dataset & \coarse\ & \fine \\
\midrule
SQuALITY & 18.5 & \textbf{6.8} \\
PubMed & 11.8  & \textbf{7.3}  \\
PubMed + ngram block & 11.7 & \textbf{9.3}  \\
\midrule
Average & 14.0 & \textbf{7.8}   \\
\bottomrule
\end{tabular}
\end{center}
\caption{Average standard deviation of faithfulness scores across annotators on a 100-point rating scale. Lower variation means higher agreement. Overall, we find that \fine-grained annotations have higher inter-annotator agreement than \coarse-grained annotations. Note that all \fine\ units of a summary were annotated to obtain these results ($f = 1.0$ in \sectionref{sec:partial-annotation}).}
\vspace{-0.1in}
\label{tab:results-variance}
\end{table}

%Standard deviation for \coarse\ annotations in SQuALITY are higher than PubMed (18.5 vs 11.7), possibly due to the 100-point Likert scale adopted by~\citet{wang2022squality} compared to the 5-point scale used in PubMed.

\noindent \textbf{Collecting \fine\ annotations}: For both SQuALITY and PubMed, we collect \fine\ annotations on Upwork (3 annotators per \fine\ unit) for the \emph{same set} of 120 summaries evaluated using \coarse\ annotations. For SQuALITY, we hire freelancers with professional experience in English, creative writing, or education. For PubMed, we hire freelancers with prior experience analyzing biomedical articles. See \appendixref{sec:fine-grained-human-evaluation} for details of our annotator screening process, compensation, instructions, and screenshots of our annotation interface.

\vspace{0.1in}

\begin{figure*}[t!]
    \centering
    \includegraphics[width=0.48\textwidth]{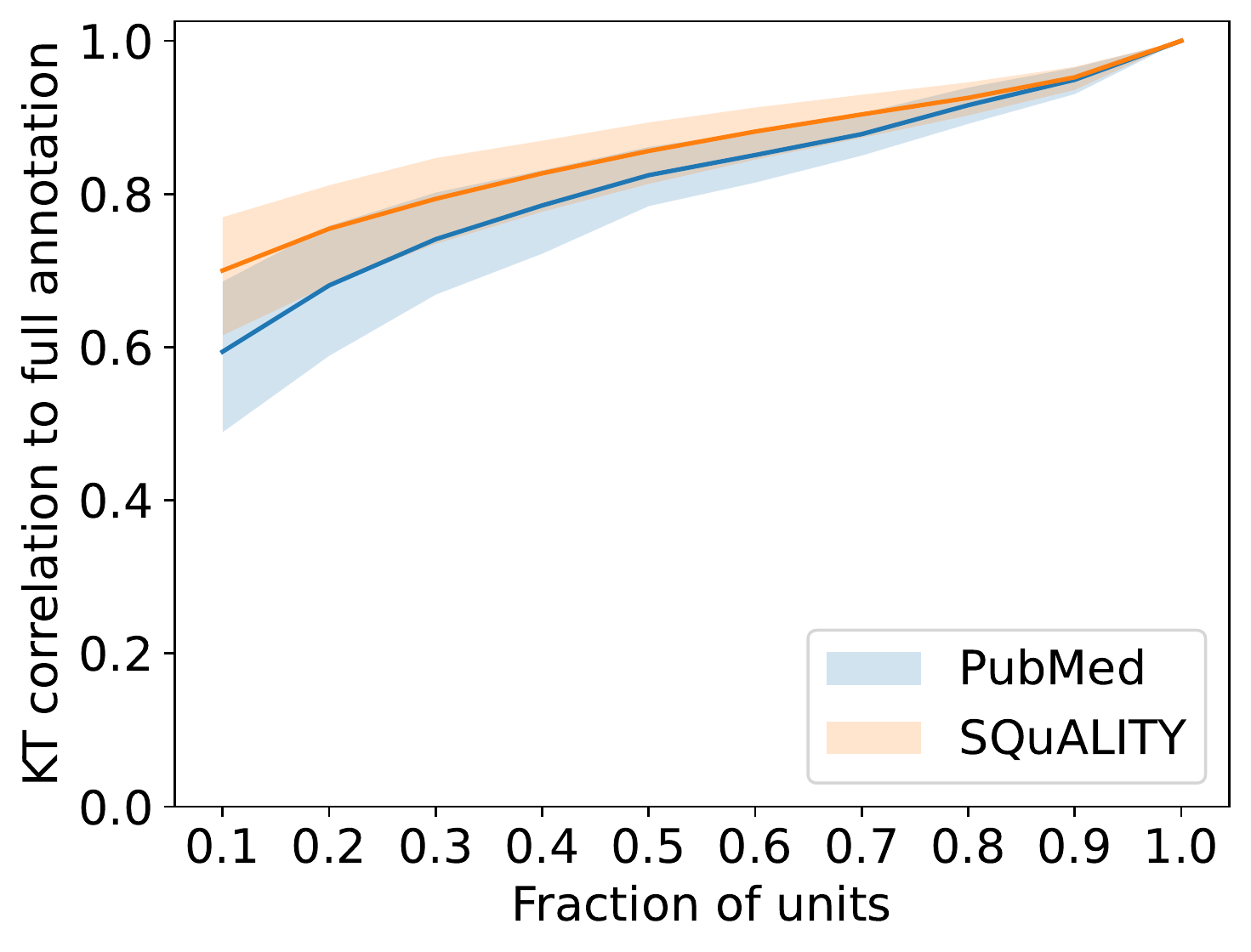}
    \includegraphics[width=0.48\textwidth]{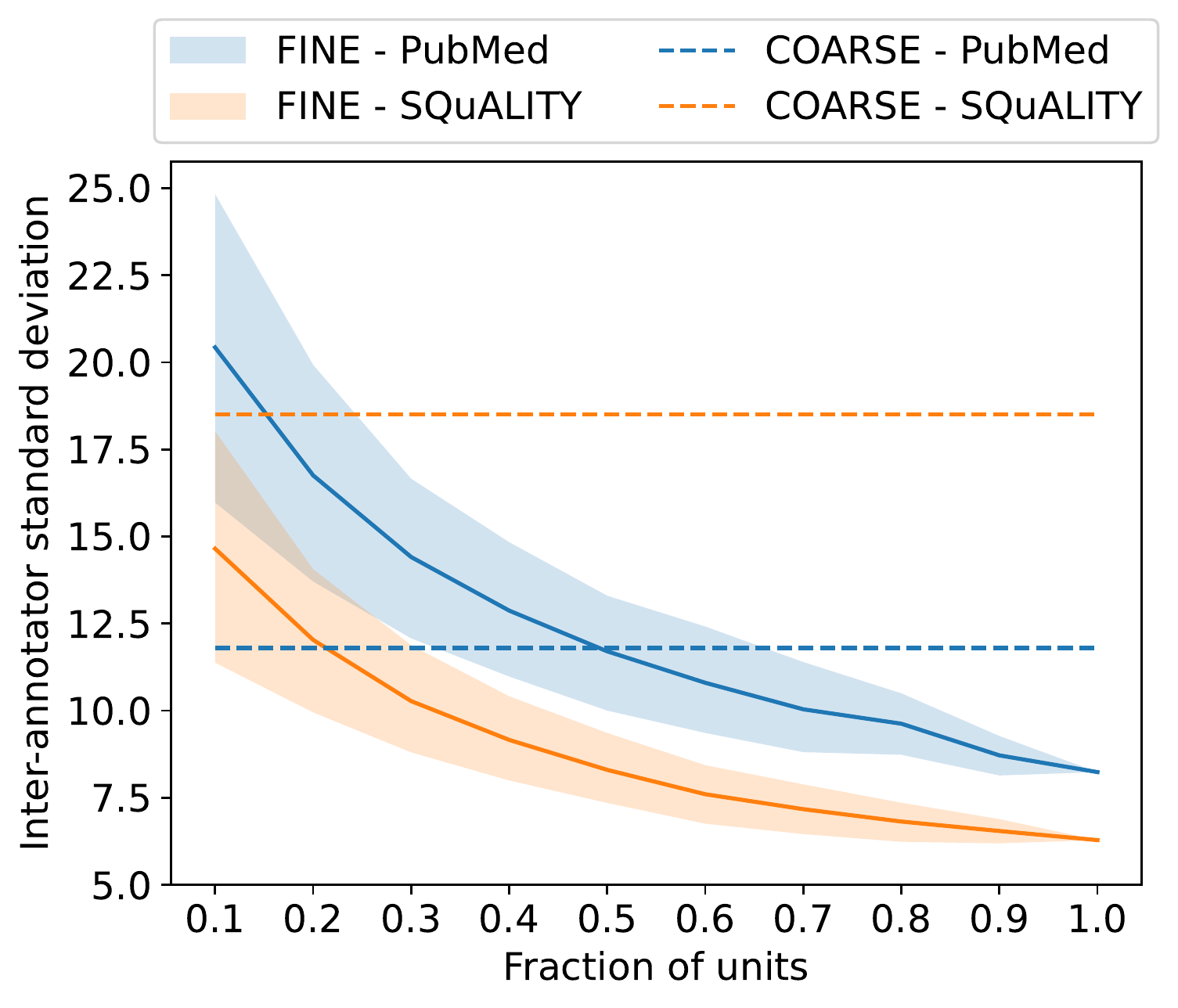}
    \caption{Accuracy and variance after annotating a fraction of units per summary (X-axis) with \fine. Despite annotating just a fraction of the summary, we observe a high segment-level Kendall tau correlation with a full annotation (left). 
    However we observe higher inter-annotator variance as the fraction reduces (right).
    % Further, we observe lower inter-annotator variance (and thus higher agreement) as the fraction annotated increases (right).
    Confidence intervals shown are 95\% and computed across 1000 random subsets (see \appendixref{appendix:partial-pearson} for left plot with Pearson).}
    \label{fig:corr-exhaustive-partial}
\end{figure*}

\noindent \textbf{\fine\ annotations have higher inter-annotator agreement than \coarse\ annotations. This leads to more confident downstream estimates}. We present our results in \tableref{tab:results-variance}. 
Overall, we observe that across all settings, \fine\ annotations have lower standard deviation (and thus higher agreement) in faithfulness scores than \coarse\ annotations (7.8 vs 14.0 average on 100-point scaled ratings). 
% Overall, we observe that across all settings, \coarse\ annotations have higher inter-annotator standard deviation than \fine\ annotations (14.0 vs 7.8 average on 100-point scaled ratings). 
To illustrate the importance of higher agreement, we measure its effect on two downstream statistics that human evaluation is primarily used for: (1) correlation with automatic metrics; and (2) mean system performance. We adapt the bootstrap resampling analysis\footnote{We slightly modify the algorithm in~\citet{deutsch2021statistical} for inter-annotator variance, see \appendixref{sec:bootstrap}.} of~\citet{deutsch2021statistical} to estimate confidence intervals of these two downstream statistics for \coarse\ and \fine.

In \figureref{fig:confidence-intervals-metrics}, we plot the 95\% confidence intervals of the Pearson correlation of various automatic evaluation metrics against \fine-grained and \coarse-grained human evaluation data. Across both datasets, \fine\ data leads to much narrower confidence intervals (0.15 vs 0.35 average uncertainty in Pearson correlation on PubMed) for the same number of summaries, implying higher statistical power. In \figureref{fig:confidence-intervals-means}, we observe a similar trend with mean system performance. Interestingly, both annotation methods give the same relative ordering of systems  (human > bart-dpr > bart for SQuALITY, human > longT5 > BigBird for PubMed-block), confirming the alignment of \fine\ and \coarse\ judgments on average.

\vspace{0.1in}

\noindent \textit{\textbf{Recommendation}}: Unlike the dominant trend in prior work, \fine-grained evaluations should be preferred over \coarse\ grained evaluation for long-form summaries. \fine\ annotations have lower inter-annotator variance than \coarse\ annotations and help localize model errors. In our setup we assume all \fine\ units are equally weighted while aggregating them to the final summary score. Despite this assumption, in our results we observe a consistent relative ordering of systems/metrics between \coarse\ and \fine\ annotations. Nevertheless, non-uniform weighing of units is an interesting future work direction; more in the Limitations section.

% 0.06 vs 0.17 pearson on SQuALITY
% 0.15 vs 0.35 pearson on PubMed

\subsection{RQ2: Can we reduce annotator workload by partially annotating a long summary?}
\label{sec:partial-annotation}

In \sectionref{sec:fine-vs-coarse-annotations}, we found that \fine\ annotations have lower variance than \coarse\ annotations. However, long summaries may be composed of several units (sentences or phrases) which each require \fine\ annotation. This could make \fine\ annotation very expensive for longer summaries (as also noted in our survey). What if we instead annotate a random subset of units from the summary? While this will lower annotation cost, how accurate would these partial annotations be? We explore this tradeoff by re-using the annotations collected in \sectionref{sec:fine-vs-coarse-annotations}. For every summary, we randomly sample a fraction of units $f \in \{0.1, 0.2 ... 0.9\}$ and then measure its correlation to the full set of annotations collected. Each annotator gets a different random sample of units for the same summary. In initial experiments, we found that this yielded higher accuracy than when keeping the same set of units per annotator.

\vspace{0.1in}

\noindent \textbf{Partial annotation has a high correlation to full annotation, but higher variance}: In \figureref{fig:corr-exhaustive-partial} (\emph{left}) we plot the segment level Kendall's $\tau$ correlation (relative ordering of summary scores) between a partial annotation and full annotation for different values of $f$. Overall, we observe a high correlation across different values of $f$. Despite annotating just half the summary ($f = 0.5$), in both datasets we observe a high correlation of 0.78-0.89 Kendall's $\tau$ (95\% interval) with a full annotation. Does a partial annotation preserve the variance benefits of \fine\ vs \coarse? In \figureref{fig:corr-exhaustive-partial} (\emph{right}) we plot the inter-annotator variance for different values of $f$. In both datasets we find that a partial annotation has a higher variance than a full annotation. While for all values of $f$ in SQuALITY we find that \fine\ annotations still have lower variance than \coarse, in PubMed \coarse\ has lower variance than \fine\ for $f <= 0.3$ with 95\% confidence.

\vspace{0.1in}

\noindent \textit{\textbf{Recommendation}}: Having annotators judge a random subset of units in a long-form summary is a simple way to reduce \fine\ annotation cost, and has high correlation with a full annotation. However, it increases inter-annotator variance. Annotating 50\% of the summary results in 0.78-0.89 Kendall's $\tau$ correlation, with a 30-40\% increase in standard deviation compared to full \fine\ annotation. Partial annotation may be limited in its ability to identify issues in summaries with very few errors. However, we find that this is not the case in current systems, which are abundant in faithfulness errors.

\subsection{RQ3: Is it useful to align summary units to sentences in the source document?}
\label{sec:highlight-hints}

So far, we have focused on design decisions on the summary side of evaluation. However, evaluating faithfulness requires a comparison of facts between a summary and a \emph{source document}. Long-form summaries tend to have long source documents (\tableref{tab:dataset-survey}): 3.1K words for SQuALITY and 5.1K words for PubMed. In \sectionref{sec:survey}, we found several mentioned human evaluation is challenging since annotators need to read long source documents. Some prior work has suggested highlighting spans in the source document that align with the summary~\citep{hardy-etal-2019-highres,kryscinski-etal-2020-evaluating,vig-etal-2021-summvis}
% to assist human evaluators
as shown in \figureref{fig:skimeval-diagram}. However, these efforts have exclusively focused on news summarization with relatively short source documents, like CNN/DM (804 words)~\citep{nallapati-etal-2016-abstractive} or XSUM (438 words)~\citep{narayan-etal-2018-dont}. How useful is highlighting based on alignment, or ``\emph{hints}'', when the spans are chosen from much longer documents?

\vspace{0.1in}

\noindent \textbf{What is the best highlighting algorithm?} 
% Before we study the usefulness of hints, 
We conduct a study to identify the alignment algorithm best suited for highlighting hints. We manually annotate 125 \fine\ units from human-written summaries of the SQuALITY validation split, marking the sentences best supporting them from the source document. We then test several candidate methods for linking summary units to the source document. These include token overlap methods like ROUGE~\citep{lin-2004-rouge}, retrievers~\citep{karpukhin-etal-2020-dense}, and fact verifiers~\citep{wadden-etal-2022-multivers}.  In \tableref{tab:linking-algorithms}, we find that SuperPAL~\citep{ernst-etal-2021-summary}, a weakly supervised linking algorithm, performs best (0.61 recall@3 vs the next best 0.47). To improve precision, we filter matches scoring less than 0.3 on SuperPAL, and show at most five highlights.

\vspace{0.1in}

\begin{table}[t!]
\footnotesize
\begin{center}
\begin{tabular}{ lrrr } 
 \toprule
 Algorithm & R@3 & R@5 & R@10 \\
 \midrule
 BM25~\shortcite{bm25} & 0.38 & 0.46 & 0.56 \\
 ROUGE-1~\shortcite{lin-2004-rouge} & 0.31 & 0.34 & 0.46 \\
% ROUGE-L~\shortcite{lin-2004-rouge} & 0.30 & 0.32 & 0.42 \\
 SIM~\shortcite{wieting-etal-2019-beyond} & 0.37 & 0.52 & 0.60 \\
 DPR~\shortcite{karpukhin-etal-2020-dense} & 0.29 & 0.31 & 0.41 \\
 BERTScore-DB-XL~\shortcite{zhang2020bertscore} & 0.30 & 0.37 & 0.46 \\
 SummaC-NLI~\shortcite{laban-etal-2022-summac} & 0.22 & 0.26 & 0.34 \\
 MultiVers-FEVER~\shortcite{wadden-etal-2022-multivers} & 0.47 & 0.58 & 0.71 \\
 SuperPAL~\shortcite{ernst-etal-2021-summary} & \textbf{0.61} & \textbf{0.68} & \textbf{0.77} \\
\bottomrule
\end{tabular}
\end{center}
\caption{A comparison of algorithms finding the top source document sentences for summary units in SQuALITY. R@$k$ (recall@$k$) denotes the fraction of times the gold sentence was in the top-$k$ predictions.}
\label{tab:linking-algorithms}
\end{table}

\begin{table}[t]
\small
\begin{center}
\begin{tabular}{ lrrrr } 
 \toprule
 Hints & Acc. ($\uparrow$) & Agree. ($\uparrow$) & \multicolumn{2}{c}{Time (secs) ($\downarrow$)} \\
 \cmidrule{4-5} 
 & (2-way) & (Fleiss) & All & First 5 \\
 \midrule
 None & \textbf{93}\% & \textbf{0.71} & 41.4 & 115.6 \\
 SuperPAL & 92\% & 0.64 & 48.2 & 84.6 \\
 Gold & 92\% & 0.63 & \textbf{40.4} & \textbf{60.4} \\
\bottomrule
\end{tabular}
\end{center}
\caption{Annotator performance (accuracy, agreement, median time) in detecting summary errors with different types of source document highlight hints. Overall, we see little difference across the three settings. }
\label{tab:linking-intrinsic-evaluation}
\end{table}

\begin{table*}[t!]
\footnotesize
\begin{center}
\begin{tabular}{ p{4cm}p{10.9cm} } 
\toprule
\bf Question \& TL;DR response & \bf Response Snippets \\
\midrule 
\textbf{Q:} Did you find the highlighted hints useful while making your judgment? \newline \newline \textbf{TL;DR}: 4 out of 5 annotators said \textcolor{mycolor3}{\textbf{Sometimes}}, 1 said \textcolor{mycolor1}{\textbf{Yes}}. More useful for SQuALITY, summary units copied verbatim from source, correct summaries. & {\scriptsize ``\textbf{With summaries that had poor correctness, the hints were often a mess}, and even correct spans had to be carefully checked. In summaries that were more correct, I could often just read the span and remember that it was correct, and then the \textbf{hints helped me find the right source position}, or \textbf{refresh my memory} about details.'' \newline ``They were more useful when the summary was a \textbf{near verbatim source reproduction}.'' \newline ``Yes, they were useful. Often they would highlight the exact passage needed to support the summary span.''  \newline ``In  \textbf{PubMed, they were a little more chaotic}, even for good summaries.'' \newline ``SQuALITY summaries consisted of sentences or parts of sentences taken straight from the story (wording was exactly as in the text). So hints often lead to the exact place.'' \newline ``For \textbf{SQuALITY, they were mostly accurate and helpful}. For PubMed, they were less accurate and relevant.''} \\
\midrule 
\textbf{Q:} Would the highlights have been sufficient to make judgments, or was reading the entire source document necessary?\newline \newline \textbf{TL;DR}: 3 out of 5 annotators said \textcolor{mycolor2}{\textbf{No}}, 2 said \textcolor{mycolor3}{\textbf{sometimes in SQuALITY}}. Reading the entire document was critical. & {\scriptsize ``\textbf{Reading the entire source document was very helpful} to understand the basic story plot'' \newline ``Even when the hints were relevant, sometimes \textbf{they left out information} (like character name)...'' \newline ``Initially I tried skimming ... then \textbf{concluded it's easier to read the entire document} first.'' \newline ``With SQuALITY there were cases where almost all of the highlights did not make any sense and nothing of that was even mentioned in the story. With \textbf{PubMed, it was even more difficult to find hints} that support the text'' \newline ``\textbf{Reading the entire document was essential} to understanding the whole process, the hints in isolation were not good enough. The \textbf{hints and the summary often confused similar objects}, especially when pronouns were involved, from different parts of the source. In PubMed a similar thing happened when the source discussed what other papers had done -- punctuation, acronyms, and abbreviations played a big role in providing context.''}\\
\midrule
\textbf{Q:} Did you use Ctrl+F searches in the source document while making judgments? \newline \newline \textbf{TL;DR}: 4 out of 5 annotators said \textcolor{mycolor1}{\textbf{Yes}}, 1 said \textcolor{mycolor3}{\textbf{yes only for PubMed}}. Ctrl+F helped locate synonyms, entities. &  {\scriptsize ``Yes, \textbf{all the time}. It was usually a \textbf{safer bet than using the hints}. The hints are given out of context of the whole SQuALITY story. There were a lot of problems with the PubMed hints involving \textbf{numbers}, which I often searched for. They were very rarely supported by the document, or contained wrong symbols (= instead of >).'' \newline ``Yes, mostly in cases the highlight did not support the summary unit partially or entirely.'' \newline ``I used Ctrl+F when looking for very specific words, like \textbf{names}. Searching was less helpful when it came to words that had synonyms or emotions.'' \newline ``I did Ctrl+F on keywords taken directly from the summary unit as well as \textbf{synonyms and any specific words} that I remembered from the story that could help me get to that place in the source document quickly.''}\\
\bottomrule
\end{tabular}
\end{center}
\vspace{-0.05in}
\caption{Results and snippets from our questionnaire with \fine\ annotators. Overall, annotators find hints only sometimes useful, and mention reading the entire source document along with keyword searches.}
\vspace{-0.05in}
\label{tab:highlight-question-survey}
\end{table*}

\noindent \textbf{Do highlighted hints improve summary error detection?} To answer this question, we manually perturb 50 \fine\ summary units in SQuALITY validation summaries, introducing entity errors or negations like \citet{kryscinski-etal-2020-evaluating}. We modify the summary context of the perturbed unit to ensure summaries are self-consistent. Annotators are shown 50 perturbed and 50 un-perturbed summaries, and asked to annotate whether the summary units are faithful to the source in three settings:\footnote{To prevent any bias, each annotator receives only one of these settings for a particular summary.} (1) no highlighted hints; (2) SuperPAL highlighted hints; (3) gold hints manually annotated by us. In \tableref{tab:linking-intrinsic-evaluation}, we show accuracy, inter-annotator agreement, and median time\footnote{Calculated using the method in~\citet{akoury-etal-2020-storium}.} for each setting.

% \vspace{0.1in}
% \kylel -- attempting to push this paragraph onto next page
\vspace{0.1in}

\noindent \textbf{Highlighted hints have almost no effect in evaluating long-form summaries}: Surprisingly, we observe that in all three metrics (accuracy, agreement, median time taken), scores are quite similar across the three settings. In fact, the ``no-hint'' setting scores slightly higher than the SuperPAL hint settings (93\% vs 92\% accuracy, 0.71 vs 0.64 Fleiss $\kappa$) and takes annotators less time (41.4 vs 48.2 seconds per unit). However, we find that hints helped annotate the first few units of a summary quicker (84.6 secs vs 115.6 secs per unit). We attribute our findings to a \emph{learning effect} over time. \fine\ annotation of long-form summaries requires annotation of several units for the same document - summary pair. As annotation progresses, annotators get more familiar with the contents of the source document and summary, reducing the need for hints over time. See \appendixref{sec:learning-effect-longform-summaries} for learning trajectory plots.

\vspace{0.1in}

\noindent \textbf{Questionnaire with \fine\ annotators confirm limited utility of hints}: Our evaluation so far is limited to perturbed human summaries. How effective are hints on model-generated summaries? To answer this, we ask five of our \fine\ Upwork annotators (from \sectionref{sec:fine-vs-coarse-annotations}) a set of three questions about their experiences using highlighted hints.\footnote{The \fine\ annotations in \sectionref{sec:fine-vs-coarse-annotations} were shown hints in the source document. Since hints may not be helpful, annotators were told not to solely rely on hints for annotation.} Detailed questionnaire results along with answer snippets are shown in \tableref{tab:highlight-question-survey}. Overall, annotators find hints were useful only sometimes. Hints were \emph{less useful} when (1) the summary unit was not supported in the source; (2) the summary unit was highly abstractive compared to the source; (3) pronouns, numbers, or abbreviations were involved; and (4) Pubmed summaries were annotated. Almost all annotators said it was necessary to read the entire source document before annotation to get an overall idea of the plot and resolve coreferences. Nearly all annotators used ``Ctrl+F'' searches along with hints to search for specific keywords while making judgments. This was especially true when the summary unit was incorrect, since the source document had to be thoroughly searched (beyond the hints) before confidently marking ``Incorrect''.

\vspace{0.1in}

\noindent \textit{\textbf{Recommendation}}: In contrast to recommendations in prior work, automatically highlighted hints are useful only in some specific cases of long-form summarization: mostly correct summaries, almost verbatim copied sentences. Annotators should be instructed to read the entire source document and to not rely solely on highlighted hints, since that could bias their judgments. Based on a small-scale study, we found SuperPAL~\citep{ernst-etal-2021-summary} to be the most accurate method for finding hints, but its performance (61\% recall@3) is far from ideal.

\subsection{To what extent do our findings generalize to short-form summarization?}
\label{sec:shortform-generalize}

In this work, we exclusively focus on summarization datasets with an average summary length of at least 150 words. This constraint excludes two popular benchmarks in summarization research over the last five years: CNN/DM~\citep{nallapati-etal-2016-abstractive} and XSUM~\citep{narayan-etal-2018-dont}. How relevant are our research questions (RQs) and findings for these short-form summarization benchmarks?

 On average, XSUM (24 words) and CNNDM (60 words) contain much shorter summaries than SQuALITY (237 words). XSUM outputs typically contain only 1 sentence or roughly 2-3 \fine\ units per summary. This blurs the distinction between \fine\ and \coarse\ units, which makes it less useful to study RQ1 in these short-form settings. The shorter length of outputs also implies that evaluation is less expensive and consumes less time, which makes our RQ2 less relevant. Finally, on average, XSUM (440 words) and CNNDM (800 words) also have much shorter source documents than datasets like SQuALITY (5200 words), reducing the need for alignment (the main premise for RQ3). The main motivation behind our study is that human evaluation of long-form summarization datasets like SQuALITY and PubMed is challenging and expensive due to the long length of the generated text. \textbf{Overall, our research questions and findings are more relevant for long-form summarization datasets than for short-form summarization datasets like XSUM and CNNDM}.
\section{Related Work}
\label{sec:related}

A large body of recent work has focused on new \emph{automatic} evaluation methods for summarization via NLI-based algorithms~\citep{falke-etal-2019-ranking, laban-etal-2022-summac} or QA-based algorithms~\citep{wang-etal-2020-asking,fabbri-etal-2022-qafacteval}. Our work focuses on the much less studied area of \emph{human} evaluation, the gold standard for developing automatic metrics. A notable effort in this space is the \textbf{Pyramid method}~\citep{nenkova-passonneau-2004-evaluating}, along with work improving Pyramid efficiency~\citep{shapira-etal-2019-crowdsourcing,zhang-bansal-2021-finding}. Efficient Pyramid-like protocols have been used to collect large-scale datasets human judgments~\citep{bhandari-etal-2020-evaluating, liu2022revisiting} in short-form news summarization tasks like CNN/DM. While these efforts focus on salience evaluation and assume access to multiple references, our work focuses on faithfulness and operates in a reference-free setting. Moreover, we focus on \emph{long-form} summarization tasks like SQuALITY and PubMed, which are much more challenging and expensive to evaluate.
\vspace{0.1in}

Evaluating summary faithfulness relates to \textbf{fact verification}~\citep{vlachos2014fact}, where claim sentences are checked against a large knowledge source (Wikipedia). Prior work~\citep{nakov2021automated} attempts to simplify the human fact checking process by methods like knowledge source snippets~\citep{fan-etal-2020-generating}, similar to hint highlights (\S\ref{sec:highlight-hints}). Faithfulness in summarization differs from fact verification in three ways: (1) summaries are paragraph-long and contextual compared to single sentence stand-alone claims in fact verification; (2) summaries are grounded to a source document, compared to a large knowledge source in fact verification; (3) summaries are model-generated compared to human-written claims in fact checking datasets~\citep{thorne-etal-2018-fever,wadden-etal-2020-fact}.

\vspace{0.1in}

\section{Conclusion}

 We present the \methodname\ guidelines, a set of recommendations for moving towards standardized human evaluation of long-form summarization. We empirically analyze each recommendation on two datasets. Overall, we find that (1) \fine-grained annotations have lower inter-annotator variance than \coarse-grained annotations; (2) partially annotating a summary reduces annotator workload while maintaining accuracy; (3) highlighting hints in the source document has limited usefulness for evaluating long-form summaries. As future work, we plan to conduct experiments on other aspects of summarization evaluation like salience and coherence.

\section*{Limitations}

Human evaluation is a noisy process with many \textbf{confounding variables}. Some of these variables were kept constant among experiments on a dataset, but modifying them could change the trends in the results. These include: (1) number of annotations per summary; (2) the specific annotation interface used; (3) granularity for \fine\ evaluation (sentences vs phrases); (4) Number of points in the Likert scale for \coarse\ evaluation; (5) set of summarization systems evaluated; and finally (6) relative (eg: A/B tests) vs absolute evaluation (eg: Likert), which has been discussed in~\citet{tang-etal-2022-investigating} for short-form news summarization datasets like CNN/DM.
\vspace{0.1in}

 Our paper is \textbf{limited to faithfulness evaluation}, but summaries are typically evaluated for salience, fluency, coherence as well~\citep{fabbri2021summeval}. While fluency may be less of an issue due to large-scale language model pretraining~\citep{dou2021scarecrow}, coherence and salience are important aspects to evaluate especially in long-form summarization~\citep{goyal2022snac}. Our findings may not generalize to evaluation of coherence or salience.

\vspace{0.1in}

Our experiments in \sectionref{sec:fine-vs-coarse-annotations} \textbf{assigned an equal weight} to each \fine\ unit while calculating the overall score of the summary. However, the faithfulness of some \fine\ units may be more important than others. A non-uniform weighing of \fine\ units may be a good strategy if there is a notion of how critical a particular unit is for a summary’s correctness. For example: (1) PICO units are critical in medical summaries~\citep{deyoung-etal-2021-ms}; (2) the Pyramid scheme~\citep{nenkova-passonneau-2004-evaluating} uses a reference frequency-based unit importance, assuming access to multiple gold references. However, a consistent notion of importance is difficult to establish across different domains, and also depends on an individual consumer’s preferences. Designing non-uniform weighing schemes is an interesting direction for future research.
\section*{Ethical Considerations}

All experiments involving human evaluation in this paper were exempt under institutional IRB review. We fairly compensated each Upwork freelancer involved in this study, at a rate of 15-20\$ per hour (respecting their suggested Upwork hourly wage). For each round of annotation, we estimated the average amount of time the task would take (by running pilots among ourselves), and provided annotators with the estimated time requirement. Most freelancers finished the task within the time window, but sometimes exceeded it by 0.5-1 hr. We compensated freelancers based on the actual time they took and their hourly wage, rather than a fixed amount per annotation.

\section*{Acknowledgments}

First and foremost, we would like to thank all the nine Upwork freelancers who contributed human annotations to this project. We are very grateful to Yixiao Song, Alex Wang, John Giorgi, Dustin Wright, Yulia Otmakhova, Daniel Deutsch, Arie Cattan, Shiyue Zhang, Tanya Goyal, Greg Durrett, Marzena Karpinska, Ankita Gupta, Nader Akoury and the Semantic Scholar team for several useful discussions at various points during the project. This work was mostly done while Kalpesh Krishna (KK) was an intern at the Allen Institute for Artificial Intelligence. KK was partly supported by a Google PhD Fellowship awarded in 2021.

\paragraph{Author Contributions:} 
\label{sec:contrib} Kalpesh Krishna led the project and performed all the technical contributions including literature review, dataset collection and processing, model implementation, annotation interface development, running experiments, and data analysis. Kalpesh also contributed to project scoping and ideation and led the writing of the paper. Erin Bransom and Bailey Kuehl helped with obtaining human judgements, including piloting the task and giving feedback,  performing the annotation themselves, and hiring and managing annotators on Upwork. Pradeep Dasigi, Arman Cohan, and Kyle Lo were mentors of the project during and after Kalpesh's internship, contributing equally to project scoping,  experimental design, ideation and direction throughout the course of the project and paper writing. Mohit Iyyer provided mentorship after the internship, in particular providing important feedback and direction on data analysis and contributing to paper writing.

% Entries for the entire Anthology, followed by custom entries
\bibliography{bib/anthology,bib/custom}
\bibliographystyle{acl_natbib}

\newpage

\appendix

\section*{Appendix}
\label{sec:appendix}

\section{Bootstrap analysis of inter-annotator variance}
\label{sec:bootstrap}

We utilize the bootstrap resampling~\citep{tibshirani1993introduction} technique described in~\citet{deutsch2021statistical} to estimate confidence intervals for human evaluation data. At a high level, bootstrap resampling helps capture the uncertainty in a downstream test statistic by repeatedly sampling from the data with replacement. We consider two downstream test statistics in our work --- (1) average system level performance; (2) correlation of human judgements to automatic metrics.

While~\citet{deutsch2021statistical} were primarily interested in uncertainty due to the specific instances and systems evaluated, our goal is to capture uncertainty due to the inter-annotator variance. Hence unlike~\citet{deutsch2021statistical}, we sample with replacement from the set of \emph{annotators} for every instance. Our precise formulation can be found in \algorithmref{alg:ci}, which operates on a $X \in \mathbb{R}^{N\times M}$ matrix of human annotations where $N$ is the number of summaries, and $M$ the number of annotators.

\begin{algorithm}[h]
{
\small
\caption{Bootstrap Confidence Interval}
\label{alg:ci}
\hspace*{\algorithmicindent} \textbf{Input:} $X \in \mathbb{R}^{N\times M}$, $k \in \mathbb{N}, \alpha \in [0, 1]$.\\
\hspace*{\algorithmicindent} ~~~~~~~~~~~~$N$ is summaries, $M$ is annotators \\
\hspace*{\algorithmicindent} \textbf{Output:} $(1-\alpha)\times 100\%$-confidence interval
\begin{algorithmic}[1]
\State samples $\gets$ an empty list
\For{$k$ iterations}
    \State $X_s \gets$ empty $N \times M$ matrix
    \For{$i \in \{1, \dots, N\}$}
    
    \State $D$ $\gets$ samp. $\{1,\dots, M\}$ w/ repl. $M$ times
    \For{$j \in  \{1, \dots M\}$}
        \State $X_s[i, j] \gets X[i, D[j]]$
    \EndFor
    \EndFor
    \State Calculate test statistic on $X_s$ and append to samples
\EndFor
\State $\ell, u \gets (\alpha/2)\times 100$ and $(1-\alpha/2)\times 100$ percentiles of samples
\State \Return $\ell, u$
\end{algorithmic}
}
\end{algorithm}

\section{Human evaluation details}
\label{sec:human-evaluation-details}

\subsection{\fine-grained evaluations of SQuALITY and PubMed summaries}
\label{sec:fine-grained-human-evaluation}

We interviewed a total of 9 Upwork freelancers for the position, offering a compensation of \$15-16.5 / hr (depending on their Upwork hourly rate). The screening procedure involved a qualification task on synthetically perturbed summaries from the SQuALITY dataset validation split. Similar to the final annotation task, annotators were shown a highlighted clause from the summary, and asked to mark whether or not it is supported by the source document. 50\% of the clauses were synthetically perturbed (via negation or entity swapping as in~\citealp{kryscinski-etal-2020-evaluating}) and manually checked to ensure they were not supported by the source document. A total of 6 freelancers scored 85\% or better, and were recruited for the main set of experiments. All 9 freelancers were compensated for the screening round at the rate of 15\$ USD / hr.

\begin{table}[t]
\begin{center}
\begin{tabular}{ lrrrr } 
 \toprule
 %Model & vs formal vec & vs informal vec \\
 & F-$\kappa$ & R-$\kappa$ & all agree  \\
 \midrule
 Random & 0.00 & 0.00 & 25\% \\
 SQuALITY & 0.74 & 0.76 & 82\%\\
 PubMed & 0.53 & 0.65 & 74\% \\
\bottomrule
\end{tabular}
\end{center}
\caption{Fleiss kappa (F-$\kappa$), Randolph kappa (R-$\kappa$), and agreement scores of our \fine\ annotation per summary unit. All $\kappa$ scores are well above a random annotation baseline, indicating good agreement.}
\label{tab:kappa-fine}
\end{table}

All six hired annotators are native or bilingual English speakers. All annotators have completed a degree at the undergraduate level and three also have Masters degrees, with the most common focuses of the degrees being English/creative writing and education. The annotators’ common professional experiences include copywriting, editing, proofreading, writing, and teaching. Finally, for PubMed annotations we re-hired three annotators from the pool of six SQuALITY annotators who mentioned they had experience reading and analyzing biomedical articles. These three annotators were provided with an additional bonus of \$30 after they completed all annotations.

\begin{figure*}[t!]
    \centering
    \includegraphics[width=0.99\textwidth]{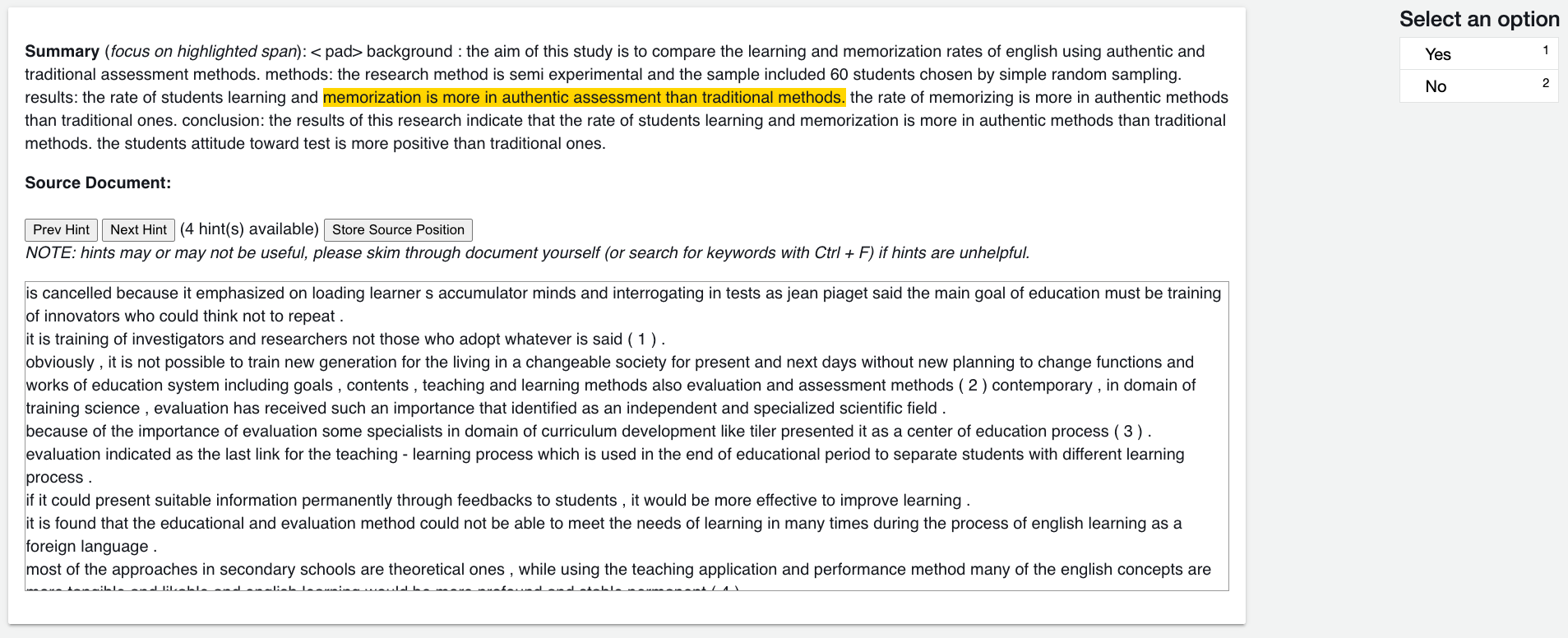}
    \caption{The AMT Sandbox annotation interface used for \fine\ evaluation of SQuALITY and PubMed summaries (\appendixref{sec:fine-grained-human-evaluation}).}
    \label{fig:skimeval-interface}
\end{figure*}

Annotators are provided with a detailed annotation guideline along with examples of faithfulness (\tableref{tab:annotation-instructions-fine}). Our guidelines are mostly consistent with a recently proposed set of guidelines for checking attribution in text generation~\citep{rashkin2021measuring}. The final annotation interface is implemented in AMT Sandbox, as shown in \figureref{fig:skimeval-interface}.

\vspace{0.1in}

\noindent \textbf{Inter-annotator agreement (binary)}: Much of the analysis in \sectionref{sec:longeval} uses standard deviation across summaries scores to measure inter-annotator agreement. However, another way to calculate inter-annotator agreement for \fine\ annotations is measuring agreement on individual units which received a Yes / No judgment. In \tableref{tab:kappa-fine} we show these inter-annotator agreement statistics. We measure Fleiss Kappa~\citep{fleiss1971measuring}, Randolph Kappa~\citep{randolph2005free,warrens2010inequalities}, and the fraction of sentence pairs with total agreement.\footnote{The $\kappa$ scores are measured using the library \url{https://github.com/statsmodels/statsmodels}.} In the table we can see all agreement statistics are well away from a uniform random annotation baseline, indicating good agreement.

\subsection{\coarse-grained evaluation of PubMed summaries}
\label{sec:coarse-grained-human-evaluation}

None of the surveyed papers evaluating PubMed summaries with humans released their human evaluation data. Hence, we decided to collect our own \coarse\ annotations. Since \fine\ annotations (\sectionref{sec:fine-grained-human-evaluation}) may have biased our original set of annotators, we hire three new annotators to perform overall assessments on a 5-point Likert scale. In other words, we use a ``between-subject'' experiment design to compare \fine\ against \coarse. 

We hired three freelancers on Upwork, all of whom have extensive professional experience reading research papers (two of them had PhDs in biomedical fields). All annotators were compensated at a rate of 20\$ USD / hr, their hourly rate on Upwork. All three annotators had been previously screened and hired by us for different projects in the past. Two of them had assisted us in an annotation task involved reading short summaries of biomedical academic papers and evaluating them for fluency, accuracy, correctness.

\begin{figure*}[t!]
    \centering
    \includegraphics[width=0.99\textwidth]{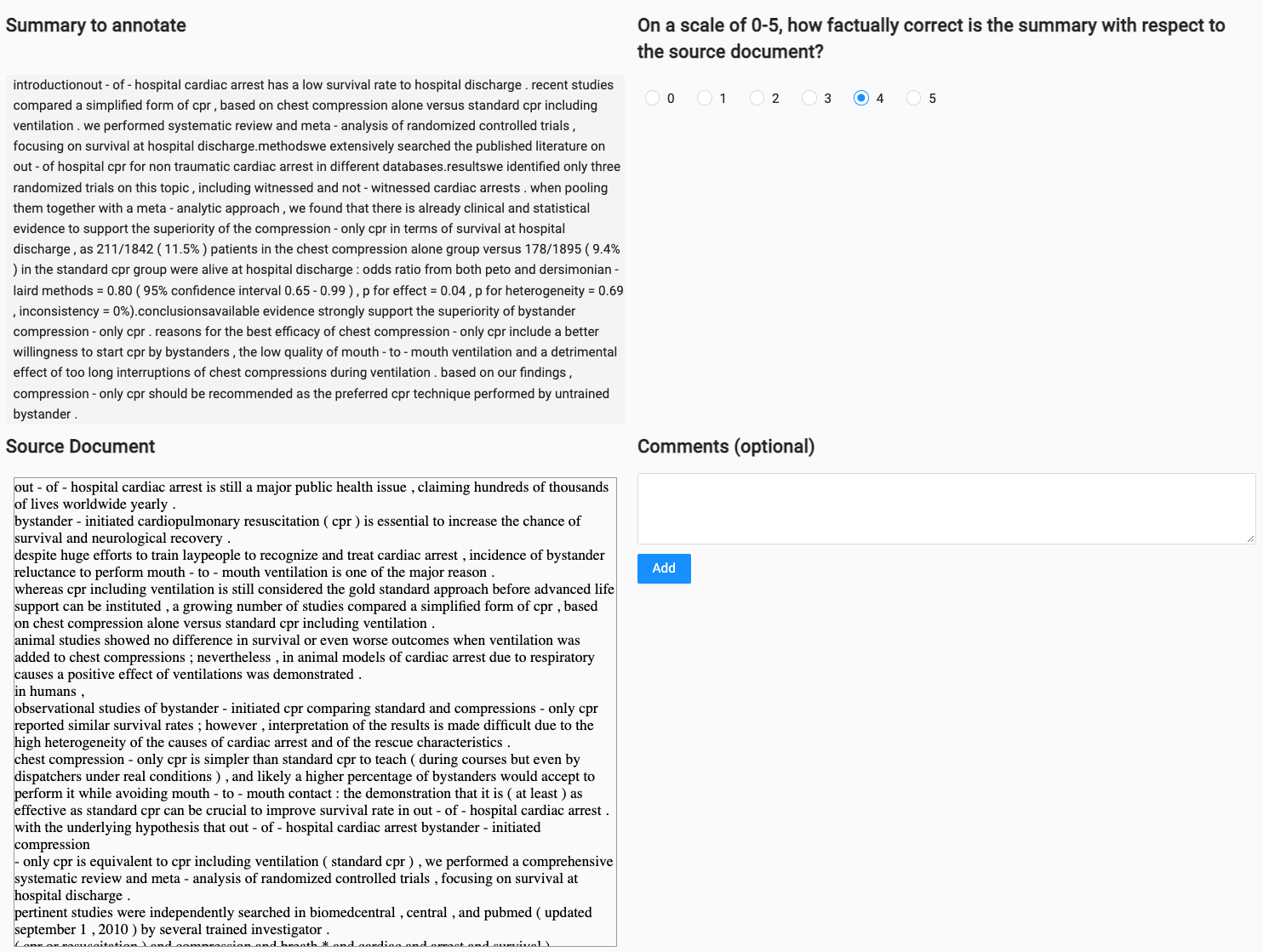}
    \caption{The LabelStudio annotation interface used for \coarse\ evaluation of PubMed summaries (\appendixref{sec:coarse-grained-human-evaluation}).}
    \label{fig:likert-interface}
\end{figure*}

Annotators are provided with a detailed annotation guideline along with examples of faithfulness (\tableref{tab:annotation-instructions-coarse}). Our guidelines are mostly consistent with a recently proposed set of guidelines for checking attribution in text generation~\citep{rashkin2021measuring}. The final annotation interface is implemented in LabelStudio, as shown in \figureref{fig:likert-interface}.

\subsection{Crowdworkers or expert annotators?}
\label{sec:annotator-choice}

Several prior works have raised the issue of low inter-annotator agreement and poor accuracy with non-expert annotators (eg: MTurk crowdworkers) in human evaluation of summarization~\citep{gillick2010non,fabbri2021summeval, falke-etal-2019-ranking} and open-ended long-form generation~\citep{karpinska-etal-2021-perils,clark-etal-2021-thats}. In our survey (\tableref{tab:results-annotators-survey}), we found the type of annotators used in long-form summarization is often not specified (16 / 43 papers). Among other papers, 10 papers use non-experts while 17 papers use expert annotators (often graduate students).

Overall, we echo the concerns with non-expert annotators and recommend hiring freelancers on Upwork (or experts) who are well-versed with the domain for annotation. In initial experiments, we attempted to recruit Amazon Mechanical Turk crowdworkers filtered by the ``Master's qualification'' and having a 90\%+ approval rating. In our qualification task of error detection in synthetically perturbed SQuALITY summaries, MTurkers scored just 62\% (binary classification) with a three-annotator Fleiss $\kappa$ of 0.15. On the other hand, Upwork freelancers (with professional writing experience) an accuracy 90\% with a high inter-annotator agreement (Fleiss $\kappa = 0.71$).

\section{Additional Survey Statistics}
\label{sec:additional-survey-stats}

In \tableref{tab:survey-best-practices} and \tableref{tab:results-annotators-survey} we document some additional statistics for the 44 papers conducting human evaluation of long-form summarization.

\begin{table}[h]
\small
\begin{center}
\begin{tabular}{ lr } 
 \toprule
 Best practice & \# papers \\
 \midrule
Raw human evaluation data released & 2 / 44 \\
Interface or instructions provided & 9 / 44 \\
 Inter-annotator agreement reported & 12 / 44 \\
Statistical analysis conducted & 12 / 44 \\
Multiple datasets are human evaluated & 14 / 44 \\
Multiple annotators per summary & 33 / 44 \\
Annotator background reported & 33 / 44 \\
Specific summary aspects evaluated & 42 / 44 \\
\bottomrule
\end{tabular}
\end{center}
\caption{Fraction of surveyed papers following the best practices recommended by~\citet{gehrmann2022repairing}. We include only the 44 papers here which conducted a human evaluation of long-form summarization.}
\label{tab:survey-best-practices}
\end{table}

\begin{table}[h]
\small
\begin{center}
\begin{tabular}{ lr } 
 \toprule
 Type of annotator & \# papers \\
 \midrule
 No details specified & 11 / 44 \\
 Native English speaker** & 5 / 44 \\
 Mechnical Turk crowdworker & 9 / 44 \\
 Non-expert volunteers & 1 / 44 \\
 \midrule 
 Extensive prior experience** & 3 / 44 \\
 Graduate students / researchers & 13 / 44 \\
 Upwork freelancers & 2 / 44 \\
\bottomrule
\end{tabular}
\end{center}
\caption{The types of annotators used across different long-form summarization papers. ** - No additional details were specified.}
\label{tab:results-annotators-survey}
\end{table}

\section{Automatic summarization metrics used for evaluation}

The following metrics are considered while measuring Pearson's correlation with our human evaluation data (\figureref{fig:confidence-intervals-metrics}) --- ROUGE-1 / 2 / F~\citep{lin-2004-rouge}, BARTScore / BARTScore-Parabank~\citep{yuan2021bartscore}, Sentence-BLEU~\citep{papineni-etal-2002-bleu}, BERTScore~\citep{zhang2020bertscore} and BLEURT~\citep{sellam-etal-2020-bleurt}. A number of metrics were calculated using the SacreROUGE repository~\citep{deutsch-roth-2020-sacrerouge}.

\section{Learning effect while annotating long-form summaries}
\label{sec:learning-effect-longform-summaries}

\begin{figure}[h]
    \centering
    \includegraphics[width=0.49\textwidth]{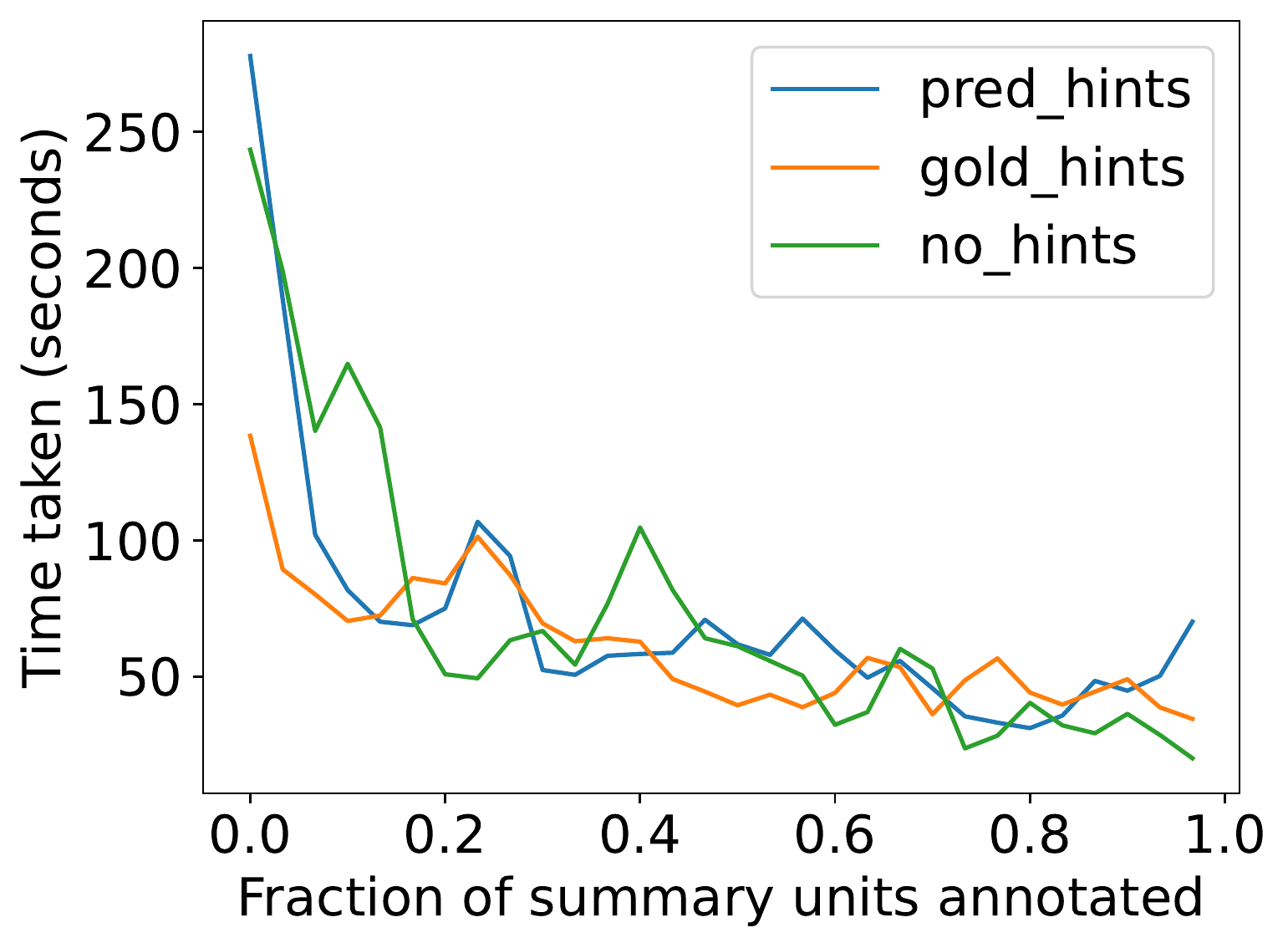}
    \caption{Learning effect over time while evaluating long-form summaries with \fine\ annotation. As the annotators evaluate more summary units, they learn the document better and are much faster at annotation irrespective of whether hints are shown to them.}
    \label{fig:learning-time}
\end{figure}

In \sectionref{sec:highlight-hints} we discussed a learning effect where annotators get more familiar with the contents of a source document as they annotate more \fine-grained units in a long-form summary. To better understand this effect, in \figureref{fig:learning-time} we plot the average time taken by annotators as they progress in their annotation of a summary. Overall, we find that annotators get significantly faster in annotating the summary after the first 20\% units. We hypothesize that annotators get pretty familiar with the general topics in the source document after the first few annotations, speeding up subsequent annotations.

\section{Partial summary annotation with pearson correlation}
\label{appendix:partial-pearson}

\begin{figure}[h]
    \centering
    \includegraphics[width=0.49\textwidth]{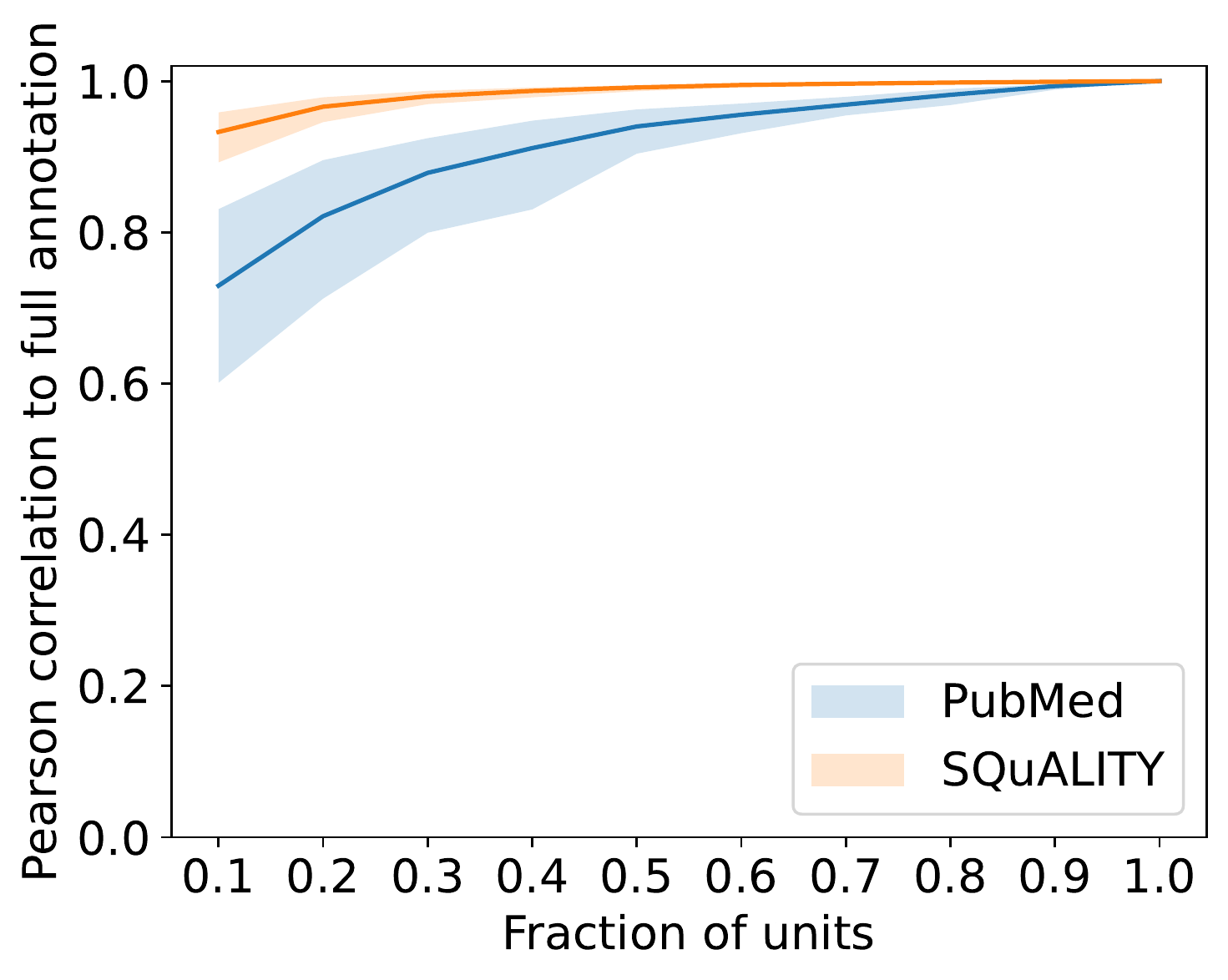}
    \caption{A version of \figureref{fig:corr-exhaustive-partial} using Pearson correlation instead of Kendall Tau correlation.}
    \label{fig:corr-exhaustive-partial-pearson}
\end{figure}

See \figureref{fig:corr-exhaustive-partial-pearson}.

\section{Metric correlations using Kendall's Tau}
\label{sec:metric-correlations-kt}

\begin{figure*}[t!]
    \centering
    \includegraphics[width=0.48\textwidth]{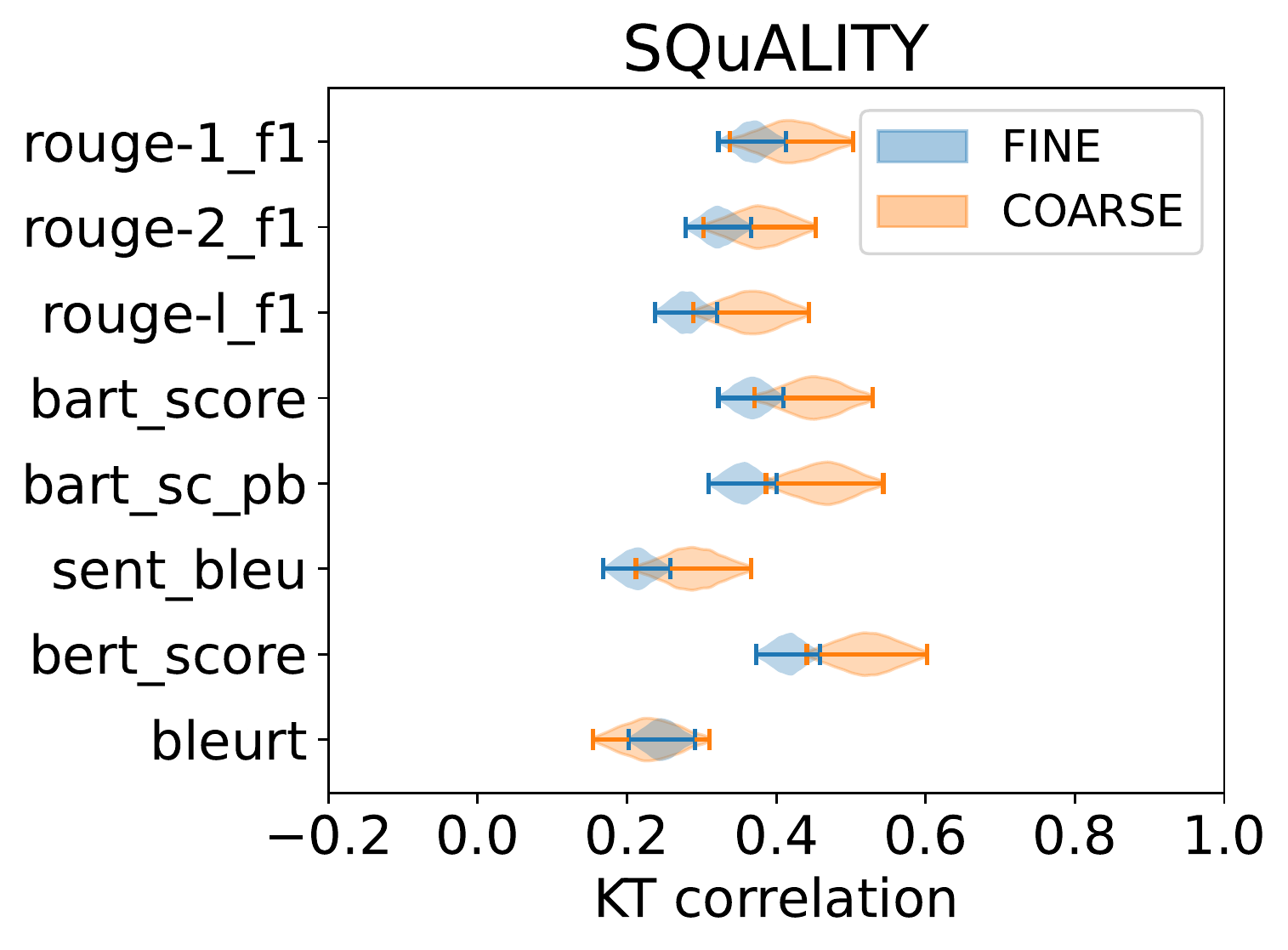}
    \includegraphics[width=0.48\textwidth]{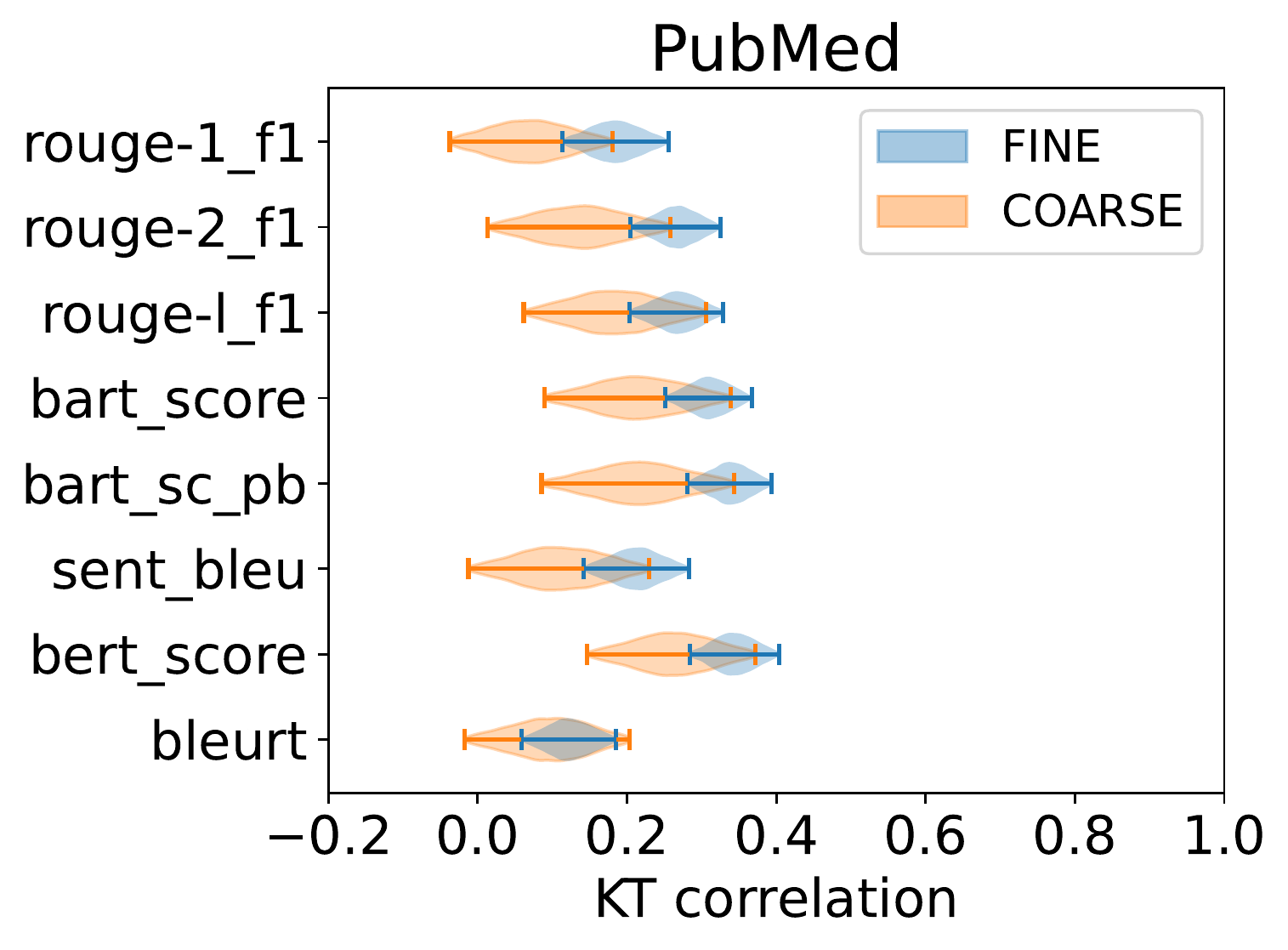}
    \caption{A version of \figureref{fig:confidence-intervals-metrics} using Kendall's Tau correlation instead of Pearson's correlation.}
    \label{fig:confidence-intervals-metrics-kt}
\end{figure*}

See \figureref{fig:confidence-intervals-metrics-kt}.

\begin{table*}[t!]
\footnotesize
\begin{center}
\begin{tabular}{ p{14.8cm} } 
\toprule
        In this task, you will be shown a long document ("Source Document") and its Summary. A span of text will be highlighted in the summary, and the goal is to check if this span is factually supported by the source document. You will need to choose one of two options: \newline 

        1. \textbf{Yes}: if \emph{\textbf{all}} the facts in the highlighted summary span are supported by the source document
        
        2. \textbf{No}: if the highlighted summary span presents some information that is not supported by the source document (either a direct contradiction, or not present)
        
        In addition to the source document, you will be provided with some highlighted text ("hints") in the source document which may help you in making a decision. Press the "Next Hint" button to scroll through the highlighted hints. Source document hints may or may not be helpful. Do not make a judgment solely based on these hints. Skim through the source document yourself / search for keywords with Ctrl + F if the hints are not helpful.
        
        Below you can find some short representative \textbf{examples}.\newline
        
        \textbf{Example 1} \newline
        Summary (only highlighted span shown) = ... Retief is not Lemuel's cousin. ... \newline
        Source Document (snippets shown) = He eyed Retief ... "He ain't no cousin of mine," Lemuel said slowly. \newline
        Supports = Yes \newline
        
        \textbf{Example 2} \newline
        Summary (only highlighted span shown) = ... Lemeul knocks down Retief. ... \newline
        Source Document (snippets shown) = Retief's left fist shot out, smacked Lemuel's face dead center. He stumbled back, blood starting from his nose; ... He caught himself, jumped for Retief ... and met a straight right that snapped him onto his back: out cold. "Wow!" said Potter. "The stranger took Lem ... in two punches!" \newline
        Supports = No (Reason: Retief knocks down Lemeul, not the other way around.) \newline
        
        \textbf{Example 3} \newline
        Summary (only highlighted span shown) = ... Potter and his team do not trust the Embassy. ... \newline
        Source Document (snippets shown) = Lemme up. My name's Potter. Sorry 'bout that. I figured it was a Flap-jack boat; looks just like 'em . He waved a hand toward the north, where the desert lay. \newline
        Supports = No (Reason: The claim is irrelevant to the evidence.) \newline \\
\bottomrule
\end{tabular}
\end{center}
\caption{Annotation guidelines provided to annotators for \fine-grained evaluation of SQuALITY and PubMed summaries. (\appendixref{sec:fine-grained-human-evaluation}).}
\label{tab:annotation-instructions-fine}
\end{table*}

\begin{table*}[t!]
\footnotesize
\begin{center}
\begin{tabular}{ p{14.8cm} } 
\toprule
Instructions for Likert-scale evaluation. Please read all instructions before starting the annotation.\newline

\textbf{Setup}

1. Start by signing up on Label Studio, you will need to provide an email ID and password. It’s okay to use a non-existent throw-away email ID here. Also, do not use any personal / sensitive passwords (but make sure to remember your email / password for logging in next time!). Click on the box saying “<your name> — Summarization Evaluation”

2. In this batch a total of 30 summaries need to be evaluated. Every three consecutive rows are different summaries of the same source document. You can evaluate a summary by clicking on a row, and annotating it. Optionally, you can click on “Label All Tasks” at the top of the screen.\newline

\textbf{Annotation Task}

Each summary needs to be evaluated for its “correctness”. You need to provide a 0-5 judgment for the entire summary, where “correctness” can be defined as, “The absence of factual errors in the summary, where a factual error is a statement that contradicts the source document, or is not directly stated, heavily implied, or logically entailed by the source document”. For example,

Source Document (snippet shown) = ….. Vitamin C was discovered in 1912, isolated in 1928, and, in 1933, was the first vitamin to be chemically produced. It is on the World Health Organization's List of Essential Medicines. Vitamin C is available as an inexpensive generic and over-the-counter medication. Partly for its discovery, Albert Szent-Györgyi and Walter Norman Haworth were awarded the 1937 Nobel Prizes in Physiology and Medicine and Chemistry, respectively. Foods containing vitamin C include citrus fruits, kiwifruit, guava, broccoli, Brussels sprouts, bell peppers, potatoes, and strawberries. Prolonged storage or cooking may reduce vitamin C content in foods. ….

Summary 1 (snippet shown) = … Chicken contains vitamin C …

Summary 2 (snippet shown) = … Albert Szent-Györgyi won the 1955 Nobel Prize for discovering Vitamin C …

Summary 3 (snippet shown) = … Vitamin C was the first chemically produced Vitamin …

Summary 4 (snippet shown) = … Apple contains vitamin C …

Errors marked in red. Here, the snippets for summary 1 are incorrect, summary 2 partially correct, and summary 3 completely correct with respect to the source document. Summary 4 is incorrect with respect to the source document (since it’s never discussed), but a globally correct fact. You should treat such a summary as incorrect since it is not mentioned in the source document.

(This is an illustrative example only, the actual annotation task has much longer summaries / source documents.)

The rating scale is from 0 to 5, where 0 is the lowest possible rating (most or all of the summary is wrong / irrelevant to the source document), and 5 is the highest rating (most or all of the summary is correct).

While it is compulsory to provide a judgment from 0 to 5 for each summary, you can optionally provide additional comments in your annotation. For instance, if the judgment needs to be more nuanced than a 5-point scale, you prefer to mark something like “3.5”, or you would like to add some other notes about your judgment.

Press “Submit” after you have provided your annotation.\newline

\textbf{Suggested workflow}

Every three consecutive rows contain different summaries for the same source document. We suggest the following workflow while annotating documents —

1. Spend the first 15 minutes reading the source document and getting a general sense of the facts mentioned in the document.

2. Spend 5 minutes to read and annotate the summaries in each of the three consecutive rows which correspond to the same document. Add optional comments / notes if necessary.

3. In the last 5 minutes, re-calibrate your ratings across the three rows if needed (for instance, you significantly preferred the correctness of summary 1 vs summary 2, but you gave it the same rating in the initial pass). Add optional comments / notes if necessary.

Following this workflow, it should take 35 minutes to annotate each set of 3 rows. For 30 rows, this should take ~6 hrs.\\

\bottomrule
\end{tabular}
\end{center}

\caption{Annotation guidelines provided to annotators for \coarse\ evaluation of PubMed summaries (\appendixref{sec:coarse-grained-human-evaluation}).}

\label{tab:annotation-instructions-coarse}
\end{table*}

\end{document}